\documentclass[10pt,journal]{IEEEtran}
\ifCLASSINFOpdf
\else
\fi
\usepackage{graphicx}
\usepackage{amssymb}
\usepackage{amsmath}
\usepackage{multirow}
\usepackage{amsmath}
\usepackage{amssymb}
\usepackage{booktabs}
\usepackage[colorlinks,linkcolor=black, urlcolor=black,citecolor=black]{hyperref}
\begin{document}
\title{Context-Integrated and Feature-Refined Network\\ for Lightweight Object Parsing}

\author{Bin Jiang,~\IEEEmembership{Member,~IEEE,}
        Wenxuan Tu,
        Chao Yang,~\IEEEmembership{Member,~IEEE,}\\
        and Junsong Yuan,~\IEEEmembership{Senior Member,~IEEE}
\thanks{B. JIANG (corresponding author) is with the College of Computer Science and Electronic Engineering, Hunan University, Changsha 410082, China (e-mail: jiangbin@hnu.edu.cn).}% <-this % stops a space
\thanks{W. TU is with the College of Computer Science and Electronic Engineering, Hunan University, Changsha 410082, China (e-mail: twx@hnu.edu.cn).}% <-this % stops a space
\thanks{C. YANG is with the College of Computer Science and Electronic Engineering, Hunan University, Changsha 410082, China (e-mail: yangchaoedu@hnu.edu.cn).}% <-this % stops a space
\thanks{J. YUAN is with the Department of Computer Science and Engineering, State University of New York at Buffalo, Buffalo, NY 14260-2500 USA (e-mail: jsyuan@buffalo.edu).}% <-this % stops a space
}
\maketitle

\begin{abstract}
Semantic segmentation for lightweight object parsing is a very challenging task, because both accuracy and efficiency (e.g., execution speed, memory footprint or computational complexity) should all be taken into account. However, most previous works pay too much attention to one-sided perspective, either accuracy or speed, and ignore others, which poses a great limitation to actual demands of intelligent devices. To tackle this dilemma, we propose a novel lightweight architecture named Context-Integrated and Feature-Refined Network (CIFReNet). The core components of CIFReNet are the Long-skip Refinement Module (LRM) and the Multi-scale Context Integration Module (MCIM). The LRM is designed to ease the propagation of spatial information between low-level and high-level stages. Furthermore, channel attention mechanism is introduced into the process of long-skip learning to boost the quality of low-level feature refinement. Meanwhile, the MCIM consists of three cascaded Dense Semantic Pyramid (DSP) blocks with image-level features, which is presented to encode multiple context information and enlarge the field of view. Specifically, the proposed DSP block exploits a dense feature sampling strategy to enhance the information representations without significantly increasing the computation cost. Comprehensive experiments are conducted on three benchmark datasets for object parsing including Cityscapes, CamVid, and Helen. As indicated, the proposed method reaches a better trade-off between accuracy and efficiency compared with the other state-of-the-art methods.

\end{abstract}

\begin{IEEEkeywords}
Object Parsing, Semantic Segmentation, Model Efficiency, Feature Refinement, Multi-scale Context Information.
\end{IEEEkeywords}

% \fi
\IEEEpeerreviewmaketitle

\section{Introduction}
\begin{figure}[!t]
\centering
\includegraphics[width=3.2in]{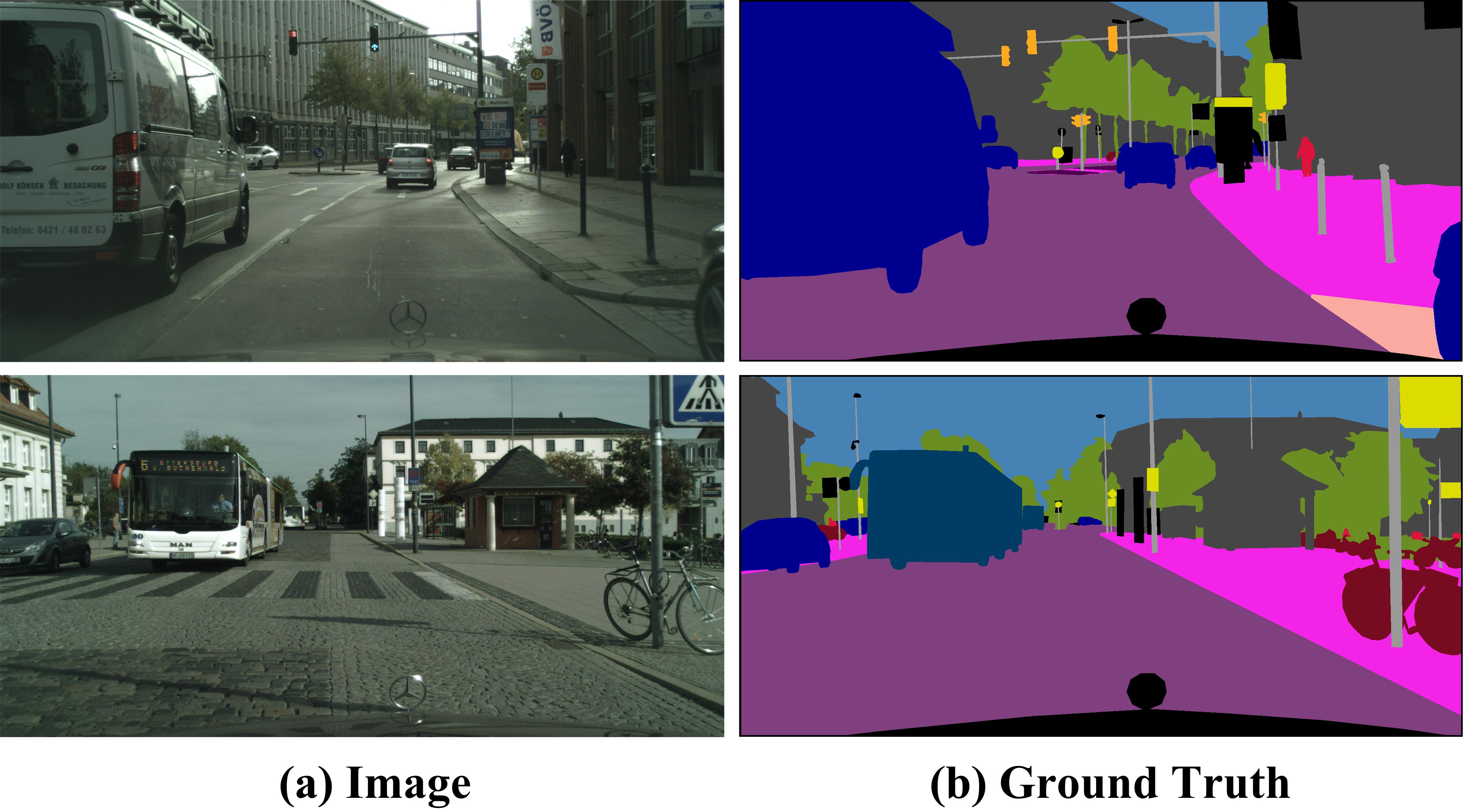}
% where an .eps filename suffix will be assumed under latex,
% and a .pdf suffix will be assumed for pdflatex; or what has been declared
% via \DeclareGraphicsExtensions.
\caption{Illustration of some urban scenes in Cityscapes dataset. An intelligent device must comprehensively implement scene interactivity with the complex environment, then make the driving decision timely.}
\label{1}
\end{figure}

\IEEEPARstart{S}{emantic} segmentation aims at labeling pixel-level signals associated with category, location, and shape for objects, which can be utilized in many applications such as robotic system, medical imaging, and object parsing \cite{2}-\cite{7}. Object parsing is supposed to segment the whole image into different semantic parts such as a building, a car, and a person, as shown in Fig. \ref{1}. The core challenge of object parsing is how to keep both high accuracy and real-time speed under resource-constrained environments. Hence for methods to be practically applicable, it is necessary that they have to be accurate, fast as well as resource-saving.

\begin{figure*}[!t]
\centering
\includegraphics[width=6.7in]{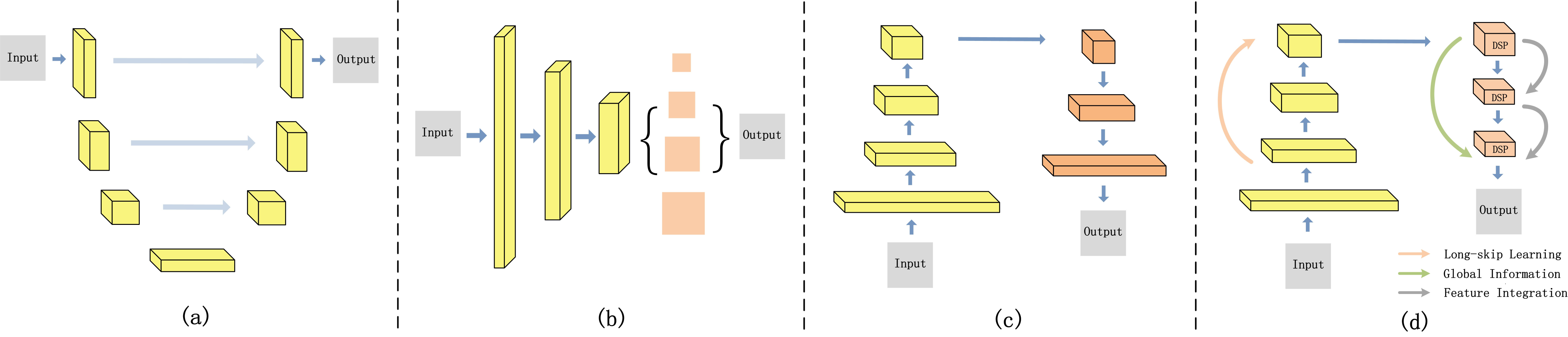}
% where an .eps filename suffix will be assumed under latex,
% and a .pdf suffix will be assumed for pdflatex; or what has been declared
% via \DeclareGraphicsExtensions.
\caption{Architecture comparison. From left to right: (a) U-shape structure. (b) Multi-scale context structure. (c) Asymmetric encoder-decoder structure. (d) The proposed CIFReNet.}
\label{2}
\end{figure*}

Early Convolutional Neural Networks (CNNs) based works handle the object parsing task by designing U-shape \cite{8}-\cite{10} or Multi-scale \cite{4}-\cite{6} architectures, which can make full use of spatial details or context information for achieving high accuracy, as shown in Fig. \ref{2}(a) and Fig. \ref{2}(b). However, these methods rely on more convolutions and sophisticated operations, which are time-consuming and require enormous computation resources. For example, although the process of details recovery could benefit from the skip learning strategy in U-shape architecture, each layer in encoder transfers an equivalent number of feature maps to corresponding layer in decoder, leading to large amounts of extra computations \cite{7}-\cite{10}. Moreover, encoding multi-scale context information on the tail of the ResNet101-based network could deal with object variation cases and boost the recognition performance, but at the expense of calculating 2,048 feature maps by 3 $\times$ 3 regular convolution before feature integration \cite{5}-\cite{6}. To overcome these drawbacks, a series of works have tried to design lightweight models to achieve a real-time speed \cite{14}-\cite{17}. As illustrated in Fig. \ref{2}(c), a typical asymmetric encoder-decoder structure focuses on reducing the number of parameters for acceleration. Nevertheless, most of them heavily compromise accuracy to speed by compressing feature channels, resulting in the final MIoU (Mean Intersection over Union) score notably drops to 60\% or even lower \cite{16}, \cite{17}.

Recently, there has been increasing interest in jointly considering a good trade-off between accuracy and speed \cite{18}-\cite{34}. Though these state-of-the-art methods could generate accurate results and maintain a real-time speed, we argue that most of them are still limited in terms of memory footprint or computational complexity. For instance, ERFNet \cite{23} employs factorized convolutions with different dilation rates in encoder to maintain the accuracy-speed balance. Unfortunately, multiple regular transposed convolution layers in decoder may suffer from amounts of invalid floating point operations. Besides, ERFNet only handles single scale cases due to the fixed-size field of view in each layer, which ignores the significance of multi-scale context information and spatial details. In order to address these issues, another strategy \cite{24} follows the principle of multi-branch framework to process multi-resolution inputs, then gathers them by a feature integration module. However, the additional branches based on regular CNNs bring large amounts of redundant parameters. Therefore, it is important to design a network that is able to perform high segmentation accuracy, and simultaneously process signals at a real-time speed under resource-constrained environments.

This motivates us to propose a lightweight architecture called Context-Integrated and Feature-Refined Network (CIFReNet), which is specifically tailored for resource constrained environments, as seen in Fig. \ref{2}(d). CIFReNet mainly includes two components: Long-skip Refinement Module (LRM) and Multi-scale Context Integration Module (MCIM). In pursuit of better accuracy, we firstly establish a single long-skip connection between shallow layers and deep layers to ease the propagation of low-frequency information. Then we apply channel attention mechanism to narrow the gap between multi-level information, which could adaptively revise some crucial details for feature refinement. After that, the Dense Semantic Pyramid (DSP) blocks, as the basic units of MCIM, are elaborately designed to aggregate multi-perspective context information together at various dilation rates. Meanwhile, the global prior knowledge is added into the MCIM to enlarge the field of view. Thus, MCIM learns the joint information of both local and global context, which guides the learning process more precisely. In order to improve the model efficiency, CIRFeNet employs the modified MobileNet V2 as encoder that contains 17 depth-wise separate convolution layers \cite{25}. The maximum number of each layer is no more than 320, which maintains the computation efficiency and guarantees adequate information during propagation simultaneously. To further reduce redundant parameters and invalid floating point operations, the depth-wise separable convolution \cite{27} and the group convolution \cite{28} are utilized to optimize the LRM and the MCIM. Note that DSP blocks are stacked in cascade rather than in parallel, refraining from large amounts of channel  calculations after feature integration. Finally, the prediction maps are bilinearly up-sampled without any additional branch or transposed convolution layer, thus improving the resource utilization ratio. We demonstrate that the proposed CIFReNet obtains 70.9\% MIoU on Cityscapes test set, 64.5\% MIoU on CamVid test set, and 71.3\% MIoU on Helen test set with less than 1.9 M parameters and only 7.3 GFLOPs. Meanwhile, it processes an image of 640 $\times$ 360 resolution at a speed of 62.5 FPS on a single NVIDIA GTX 1080Ti card. Experimental results demonstrate that our method reaches a better trade-off among overall performance compared with some state-of-the-art ones.

The key contributions of this paper are three-fold:
\begin{itemize}
\item A lightweight architecture named CIFReNet is proposed for object parsing. Compared with some state-of-the-art methods, CIFReNet obtains a better trade-off between accuracy and efficiency (e.g., execution speed, memory footprint, or computational complexity). As a side contribution, we will release our source \href{https://github.com/WxTu}{code}$\footnote{https://github.com/WxTu}$ .
\item A slight yet effective LRM is designed, which adopts a long-skip connection with channel attention mechanism to provide a highway and a proper guidance for spatial information learning. Therefore, it can adaptively refine segmentation results in a coarse level for better accuracy.
\item An efficient and powerful MCIM with cascaded DSP blocks is presented to capture multi-perspective context information and expand the field of view. In particular, the DSP block can encode much denser semantic information with an acceptable cost.
\end{itemize}

The remainder of this paper is organized as follows. Section II reviews related works in terms of object parsing tasks based on sematic segmentation. Section III presents the model design and each component of CIFReNet. Section IV conducts experiments and discusses the results. Finally, section V draws a conclusion.

\section{Related Works}
Some recent works based on Fully Convolution Networks (FCNs) \cite{29} have achieved promising results on public benchmarks \cite{31}, \cite{32}. We then review the latest deep-learning-based methods for object parsing from lightweight-oriented and accuracy-oriented aspects.

\subsection{Lightweight-oriented Approaches}
Lightweight semantic segmentation methods aim at solving the problem of slow speed and resource constraints on intelligent devices. These works can roughly be grouped into two categories: the speed-fast structure \cite{14}-\cite{17} and the accuracy-speed trade-off structure \cite{18}-\cite{34}. As for the former, ENet \cite{16} adopts the combination of an initial block and a group of factorized filters for acceleration. Further, ESPNet \cite{17} assembles Efficient Spatial Pyramid (ESP) modules into an encoder-decoder structure, outperforming ENet in terms of accuracy and speed. Although efforts have been made on designing an extremely speed-fast and resource-saving model, most of these methods sacrifice too much accuracy, so they can not generate sufficient precise information. The latter category tries to balance accuracy and speed, which has become an active research area in the last few years. Specifically, ERFNet \cite{23} designs a convolutional factorization technique with dilation rate to make parameters less redundant and obtain the acceptable accuracy. Besides, ICNet \cite{24} designs multi-branch cascaded sub-networks to achieve a fast speed, and then applies multi-scale resolution images as inputs for coarse-to-fine inference. Despite their success, so far most of them have taken less consideration on either space constraints or computation requirements, which is unfeasible for intelligent devices that require low memory footprint and computational complexity. Different from the aforementioned works, the proposed CIFReNet enhances the model efficiency by designing lightweight yet powerful LRM and MCIM, which boosts the spatial information learning and encodes denser multiple context information respectively.
\subsection{Accuracy-oriented Approaches}
Towards high-quality results, most accuracy-oriented methods are designed to encode more spatial details and multi-scale context information.

%\IEEEpubidadjcol

\subsubsection{Feature Refinement}
Refining spatial information is a common challenge in semantic segmentation methods \cite{7}-\cite{10}, \cite{39}, \cite{42}, which plays an important role in predicting the pixel-level localization. A direct solution is to design a symmetrical encoder-decoder model. For example, U-Net \cite{7} adopts long-skip connections to refine details by fusing the hierarchical features of the backbone. Similarly, SegNet \cite{8} adopts a typical U-Shape structure and utilizes the saved pooling indices to gradually compensate the spatial information for high-level features. U-shape design has also been presented in \cite{9} and \cite{10}, the authors create an up-sampling stage corresponding to each down-sampling one, and fuse them in the decoder step by step. However, most of them directly utilize element-wise sum or channel concatenation to bridge the gap among multi-level features, resulting in both high computation burden and less efficiency of feature representations. In contrast, we adopt a simple long-skip residual learning module to keep low-frequency information easier bypassed from bottom to top. Meanwhile, we introduce more semantic information to guide the low-frequency features learning, which effectively refines the final prediction with slight computation increase.

\subsubsection{Multi-scale Context Integration}
Multiple context information is generally regarded as a key factor to provide a good descriptor in object parsing works \cite{4}-\cite{6}, \cite{50}-\cite{63}. PSPNet has exhibited the impressive performance by designing Spatial Pyramid Pooling (SPP) module for capturing abundant context information\cite{4}. DeepLab V2 integrates multi-scale context information by proposing Atrous Spatial Pyramid Pooling (ASPP) module with diverse dilation rates \cite{5}. Following the same strategy, Chen \textit{et al.} \cite{13} feed global features into local context information to obtain larger field of view. Further, Ding \textit{et al.} \cite{50} manage to selectively integrate multi-scale features for each spatial position by a scheme of gated sum. Subsequently, Zhang \textit{et al.} \cite{52} present a scale-adaptive convolution to exploit long-range context information, which acquires a flexible size of receptive field. Although these methods perform well in solving the challenge of scene variations, most of them prefer heavy backbones and complicated modules to pursue high accuracy. These defects lead to the time-consuming inference and bring much heavy overheads for object parsing. In order to reduce the computation burden while maintaining high accuracy, we propose the MCIM to aggregate both local and global context information with negligible computation cost.

\section{Context-Integrated and Feature-Refined Network}
\subsection{Overview}
In this section, we introduce a single-shot architecture called CIFReNet for lightweight object parsing, which mainly consists of the Long-skip Refinement Module (LRM) and the Multi-scale Context Integration Module (MCIM). CIFReNet aims to achieve an overall trade-off in terms of accuracy and efficiency (e.g., execution speed, memory footprint, or computational complexity) by efficiently learning spatial and contextual information.

\begin{figure*}[!t]
\centering
\includegraphics[width=6.7in]{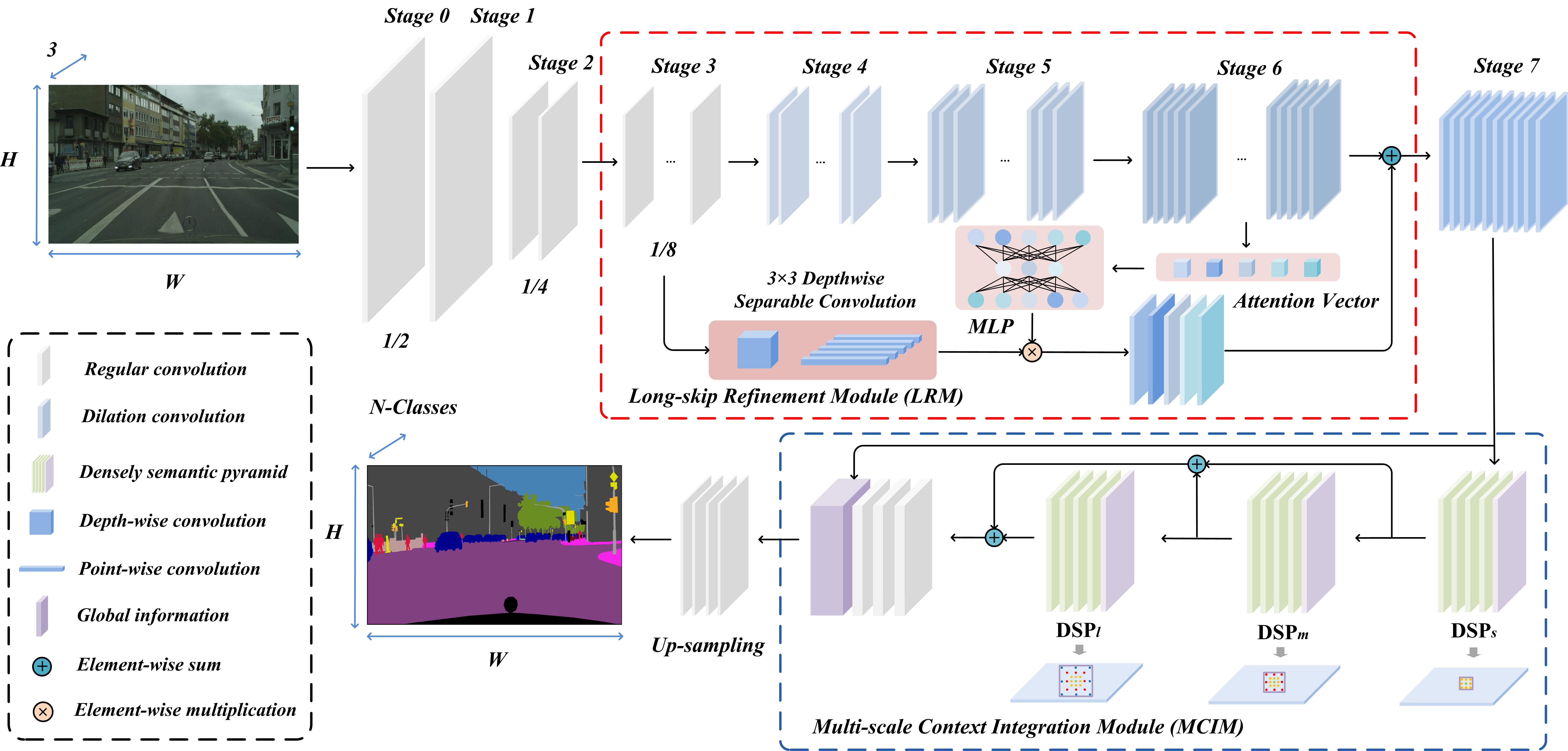}
% where an .eps filename suffix will be assumed under latex,
% and a .pdf suffix will be assumed for pdflatex; or what has been declared
% via \DeclareGraphicsExtensions.
\caption{An overview of Context-Integrated and Feature-Refined Network (CIFReNet). The dotted red
box and blue box represent the Long-skip Refinement Module (LRM) and the Multi-scale
Context Integration Module (MCIM), respectively.}
\label{3}
\end{figure*}

\begin{table}[!t]\caption{Network Architecture. \textit{C$_{o}$}: Output channels. \textit{N}: The number of categories. LRM: Long-skip Refinement Module. MCIM: Multi-scale Context Integration Module. UL: Up-sampling Layer.}
\centering
  \begin{tabular}{c|c|c|c|c}
  \toprule[1pt]
  Model&Type&Dilation&\textit{C$_{o}$}&Repeat\\
  \midrule[0.5pt]
  Stage$_{0}$&Inverted Residual Block&1&32&1\\
  Stage$_{1}$&Inverted Residual Block&1&16&1\\
  Stage$_{2}$&Inverted Residual Block&1&24&2\\
  Stage$_{3}$&Inverted Residual Block&1&32&3\\
  LRM&-&-&160&1\\
  Stage$_{4}$&Inverted Residual Block&2&64&4\\
  Stage$_{5}$&Inverted Residual Block&3&96&3\\
  Stage$_{6}$&Inverted Residual Block&5&160&3\\
  Stage$_{7}$&Inverted Residual Block&7&320&1\\
  \midrule[0.5pt]
  MCIM&-&-&400&1\\
  UL&Up-sampling Layer&-&\textit{N}&1\\
  \bottomrule[1pt]
  \end{tabular}
\label{I}
\end{table}

\begin{table}[!t]\caption{Basic notations for the proposed method.}
\centering
  \begin{tabular}{c|c}
  \toprule[1pt]
  Notations & Meaning \\
  \midrule[0.5pt]
  \textit{H $\times$ W}& The size of feature maps\\
  \textit{C$_{i}$}& The number of input channels\\
  \textit{C$_{o}$}& The number of output channels\\
  \textit{F$_{s}$}& The low-dimensional shallow feature maps\\
  \textit{F$^{'}$$_{s}$}&The high-dimensional shallow feature maps\\
  \textit{F$_{a}$}& The high-dimensional abstract feature maps\\
  \textit{V}& The set of attention vectors for \textit{F$^{'}_{s}$}\\
  \textit{V}$^{'}$& The set of dimension-reduced attention vectors\\
  \textit{I$^{H \times W \times C_{i}}$}& The input of DSP\\
  \textit{I$^{'}$$^{H \times W \times (C_{o} \times r)}$}& The channel-reduced feature maps of DSP\\
  \textit{D}& The set of dilation rates in DSP\\
  \textit{L}& The set of \textit{n}-groups features in DSP\\
  \textit{G$^{1 \times 1 \times (C_{o} \times r)}$}& The global feature vectors in DSP\\
  \textit{G$^{'}$$^{H \times W \times (C_{o} \times r)}$}& The global context in DSP\\
  \textit{O$^{H \times W \times C_{o}}$}&  The output of DSP\\
  \textit{O$^{'}$$^{H \times W \times C_{o}}$}& The residual output of DSP\\
  \textit{O$^{''}$${^{H \times W \times C_{o}}}$}& The output of the integrated semantics in MCIM\\
  \textit{F$^{H \times W \times C_{i}}_{Stage_{7}}$}& The feature maps from the tail of the encoder\\
  \textit{G$^{''}$${^{H \times W \times C}}$}& The global context generated by \textit{F$^{H \times W \times C_{i}}_{Stage_{7}}$}\\
  \textit{Y$^{''}$${^{H \times W \times (C_{o} + C)}}$}& The output of MCIM\\
  \bottomrule[1pt]
  \end{tabular}
\label{II}
\end{table}

As depicted in Fig. \ref{3}, given an input, we firstly feed it into our backbone network to obtain the semantic features. The Output Stride (OS) of the encoder is reasonably set to 8 for high-resolution datesets (e.g., Cityscapes \cite{31}) and set to 4 for low-resolution datesets (e.g., CamVid \cite{53} and Helen \cite{78}), so as to save the memory resources and preserve more spatial details during the training process. Furthermore, we replace the last four sub-sampling operations with dilation convolutions, and apply a group of hybrid dilation rates \{2, 3, 5, 7\} to maintain the field of view, as illustrated in Table \ref{I}. Note that all initial settings mentioned above are determined according to ablation studies.

Next, we perform a Long-skip Refinement Module, as shown in the red box of Fig. \ref{3}. For the first step, we establish a simple long-skip connection between spatial layer and semantic layer. The high-dimensional spatial features are generated by a 3 $\times$ 3 depth-wise separate convolution. For the second step, the high-level features are transformed into a group of weight vectors and used to refine the high-dimensional spatial features. Through the above process, we combine the refined features with original high-level features by an element-wise sum operation. Such a design eases the propagation of low-frequency information and boosts the quality of feature refinement for better segmentation results. Moreover, LRM contains only one slight long-skip connection with negligible overheads.

Thereafter, we feed semantic features outputted from the tail of the backbone into Multi-scale Context Integration Module to gather both local and global context information. As shown in the blue box of Fig. \ref{3}, three lightweight DSP blocks with a global constraint are piled up in the network. Such a cascaded design has two advantages. On the one hand, it can enlarge the field of view by increasing the depth of network. On the other hand, it follows the principle of ``deep and thin" \cite{54}, which considerably boosts the model efficiency. Additionally, the semantic features outputted from each DSP block are integrated together, which allows the network to jointly learn context information at multi-level scales for better recognition performance.

Finally, we directly up-sample the output feature maps back to input size by Up-sampling Layer (UL). In the following sections, we will elaborate the design of LRM and MCIM in detail. Before that, we summarize all notations in Table \ref{II}.

\subsection{Long-Skip Refinement Module (LRM)}
As discussed in Section I, the U-shape architecture illustrates that skip learning is beneficial to model performance improvement, but suffers from low speed and double resource utilization. Inspired by the success of long-skip learning \cite{56} and channel attention mechanism \cite{57}, we propose the Long-Skip Refinement Module to tackle the above issue for better segmentation performance by effectively and efficiently learning spatial information.

\begin{figure*}[!t]
\centering
\includegraphics[width=7.0in]{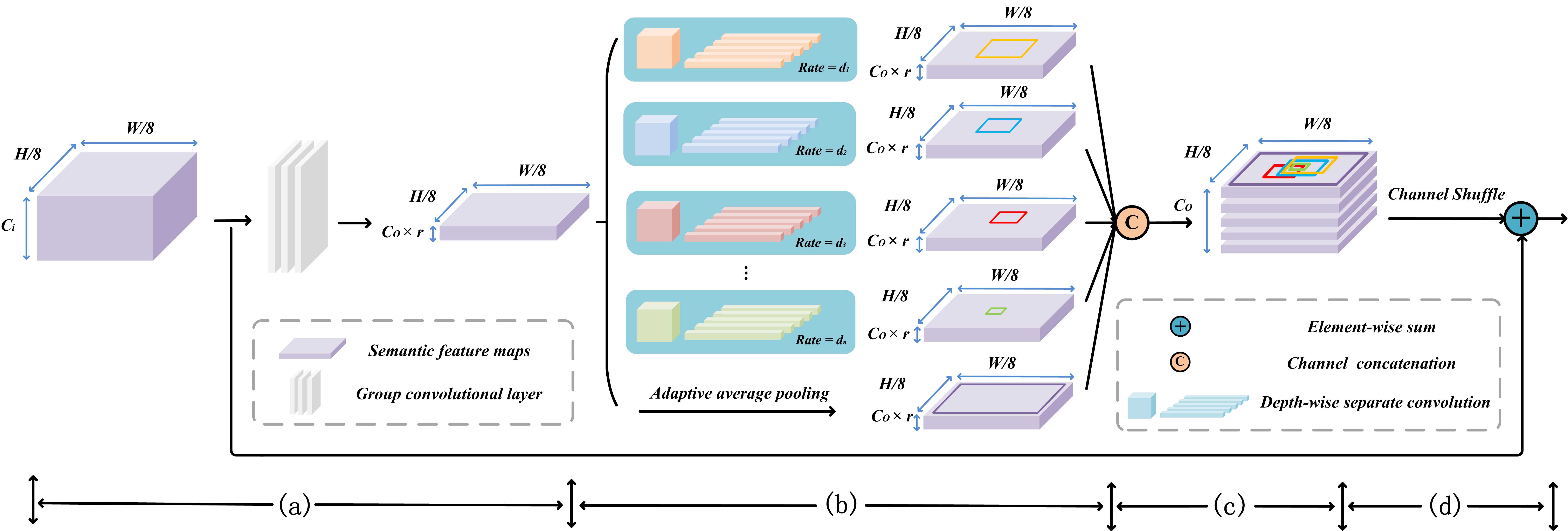}
% where an .eps filename suffix will be assumed under latex,
% and a .pdf suffix will be assumed for pdflatex; or what has been declared
% via \DeclareGraphicsExtensions.
%\captionsetup{justification=centering}
\caption{An overview of the Dense Semantic Pyramid (DSP) block. (a) Channel reduction. (b) Dense dilation sampling. (c) Semantic features integration. (d) Channel shuffle operation. }
\label{4}
\end{figure*}
\subsubsection{Basic long-skip learning} Deeper stage provides more semantic information but loses too many visual details. Although shallower stage preserves more spatial details, it contains much irrelevant noise. Based on this observation, we firstly manage to establish a long-skip connection to directly integrate multi-level features between shallow and deep stages. Extensive ablation studies have demonstrated that adopting a long-skip learning manner from Stage$_{3}$ to Stage$_{6}$ delivers a better trade-off in terms of overall performance compared with other options, which contains ten inverted residual blocks (as illustrated in Table \ref{I}) in total.

As depicted in the red box of Fig. \ref{3}, given shallow feature maps (generated by Stage$_{3}$) as an input \textit{F$_{s}$} = $\{\textit{f$_{s_{1}}^{H \times W}$, f$_{s_{2}}^{H \times W}$, $\cdots$ , f$_{s_{C_{i}}}^{H \times W}$}\}$, we firstly apply a depth-wise separable convolution to transform low-dimensional \textit{F$_{s}$} into high-dimensional \textit{F$^{'}_{s}$} = $\{\textit{f$^{'}$$_{s_{1}}^{H \times W}$, f$^{'}$$_{s_{2}}^{H \times W}$, $\cdots$ , f$^{'}$$_{s_{C_{o}}}^{H \times W}$}\}$, which is beneficial for feature representations but not computation expensive, as formulated in Eq. (1). Then we fuse \textit{F$^{'}_{s}$} and abstract features (generated by Stage$_{6}$) \textit{F$_{a}$} = $\{\textit{f$_{a_{1}}^{H \times W}$, f$_{a_{2}}^{H \times W}$, $\cdots$ , f$_{a_{C_{o}}}^{H \times W}$}\}$ by element-wise sum, which serves as the basic long-skip structure for residual learning. By this way, more abundant low-frequency information can be bypassed conveniently in the network.
\begin{equation} \label{eq:1}
F^{'}_{s} = \delta(pw^{1\times1}\times \delta(dw^{3 \times 3} \times F_{s}+b)+b),
\end{equation}
where \textit{dw$^{3\times3}$} and \textit{pw$^{1\times1}$} are defined as a 3 $\times$ 3 depth-wise convolution and a 1 $\times$ 1 point-wise convolution, respectively. \textit{b} represents the bias vector. $\delta$($\cdot$) indicates the operations of both Batch Normalization (BN) \cite{59} and Parametric Rectified Linear Unit (PReLU) \cite{60} function.

\begin{table}[!t]\caption{ Different types of convolutions and modules for comparison. $\sharp$Params: The number of parameters. \textit{m$^{2}$} is the kernel size. \textit{g} is the number of groups. \textit{n} is the number of path. RC: Regular Convolution. GC: Group Convolution. DSC: Depth-wise Separable Convolution. RC-DSP: Regular Convolution in DSP. Assume that \textit{C$_{i}$} = 320, \textit{C$_{o}$} = 320, \textit{r} = 1/4, \textit{m} = 3, \textit{g} = 4 and \textit{n} = 4.}
\centering
  \begin{tabular}{c|c}
  \toprule[1pt]
  Type&$\sharp$Params(M)\\
  \midrule[0.5pt]
  RC& \textit{C$_{i}$}$\times$\textit{C$_{o}$}$\times$\textit{m$^{2}$} = 0.92\\
  GC& (\textit{C$_{i}$}$\times$\textit{C$_{o}$}$\times$\textit{m$^{2}$})/\textit{g} = 0.23\\
  DSC& 1$\times$\textit{C$_{i}$}$\times$\textit{m$^{2}$}+\textit{C$_{i}$}$\times$\textit{C$_{o}$}$\times$1$\times$1 = 0.11\\
  \midrule[0.5pt]
  RC-DSP&(\textit{C$_{i}$}$\times$(\textit{C$_{o}$}$\times$\textit{r}))+\textit{m$^{2}$}$\times$(\textit{C$_{o}$}$\times$r)$^{2}$$\times$\textit{n} = 0.26\\
  DSP& \textit{C$_{i}$}$\times$(\textit{C$_{o}$}$\times$\textit{r}))/\textit{g}+((\textit{C$_{o}$}$\times$\textit{r})$\times$\textit{m$^{2}$}+(\textit{C$_{o}$}$\times$\textit{r})$^{2}$)$\times$\textit{n} = 0.03\\
  \bottomrule[1pt]
  \end{tabular}
\label{III}
\end{table}

\subsubsection{Spatial feature refinement} Though Zhang \textit{et al.} \cite{56} have proven that long-skip connection not only eases the optimization process but also promotes network learning in a coarse level, we argue that straightly embedding shallow features along with much noise into valuable semantic features is meaningless. To tackle this dilemma, we utilize the high-level features to provide a constraint for low-level features learning rather than simply performing element-wise sum or concatenation. Specifically, we reshape the high-level feature maps \textit{F$_a$} into feature vectors \textit{V} =  $\{\textit{v$_{1}^{1 \times 1}$, v$_{2}^{1 \times 1}$, $\cdots$, v$_{C_{o}}^{1 \times 1}$}\}$ via a global average pooling layer. Then we feed \textit{V} into Multi-Layer Perception (MLP) layers to obtain dimension-reduced feature vectors \textit{V$^{'}$} =  $\{\textit{v$^{'}$$_{1}^{1 \times 1}$, v$^{'}$$_{2}^{1 \times 1}$, $\cdots$, v$^{'}$$_{C_{o}}^{1 \times 1}$}\}$, similar to SENet \cite{57}. Finally, a softmax layer is applied to calculate the attention value which allows \textit{F$^{'}_s$} to adaptively adjust its selection. We present this process as follows:
\begin{equation} \label{eq:2}
v_{k} = \frac{1}{H \times W}\sum_{i=1}^H\sum_{j=1}^Wf_{a_{k}}(i,j),
\end{equation}
\begin{equation} \label{eq:3}
y_{k}^{H \times W} = \frac{exp(v^{'}_{k})}{\sum_{i=1}^{C_{o}}exp(v^{'}_{k})} \otimes f^{'}{_{s_{k}}^{H \times W}}\oplus f{_{a_{k}}^{H \times W}},
\end{equation}
where \textit{y}$_{k}^{H \times W}$ refers to the \textit{k}-th refined output feature map, \textit{k} $\in$ \{1, 2, $\cdots$, \textit{C$_{o}$}\}. $\otimes$ and $\oplus$ refer to element-wise multiplication and element-wise sum, respectively.
\subsection{Multi-scale Context Integration Module (MCIM)}
In the task of object parsing, most current methods design multi-scale modules such as PPM \cite{4} and ASPP \cite{5} to effectively deal with object variation cases, but heavily decrease the model efficiency. To overcome this limitation, the Multi-scale Context Integration Module that consists of three DSP blocks and global information is established, as shown in blue box of Fig. \ref{3}. The MCIM has only 0.12 M parameters and requires fewer resource utilization (0.49 MB and 2.48 GFLOPs). Next, we will further discuss about both efficiency and effectiveness of MCIM.

\subsubsection{Dense Semantic Pyramid (DSP) }
The Dense Semantic Pyramid block, with less than 0.03 M parameters, is the core component of MCIM. We then reveal the design of DSP block in detail as the following four steps.

$\textbf{Lightweight convolutional techniques.}$ As we all know, depth-wise separable convolution \cite{27} and group convolution \cite{28} have been proven more efficient in terms of memory footprint and computational complexity while keeping similar accuracy compared with the regular convolution. Therefore, we take advantages of them to make the DSP block more computationally efficient. As displayed in Fig. \ref{4}, given feature maps \textit{I$^{H \times W \times C_{i}}$} as an input of the DSP block, we firstly feed it into a 1 $\times$ 1 group point-wise convolution layer to obtain channel-reduced feature maps \textit{I$^{'}$$^{H \times W \times (C_{o} \times r)}$}, where \textit{r} refers to channel reduction ratio. Then \textit{I$^{'}$$^{H \times W \times (C_{o} \times r)}$} are successively sent to \textit{n} parallel depth-wise separable convolutions followed by BN and PReLU, in order to generate \textit{n} groups feature maps \textit{L} = $\{\textit{l$_{1}^{H \times W \times (C_{o} \times r), d}$, l$_{2}^{H \times W \times (C_{o} \times r), d}$, $\cdots$ , l$_{n}^{H \times W \times (C_{o} \times r), d}$}\}$, where \textit{d} denotes the dilation rate. With these lightweight convolutional techniques \cite{27}, \cite{28}, the dense convolutional filters among all channels are uniformly changed to be sparse, thus decreasing the computation costs and memory requirements. As listed in Table \ref{III}, the results of bottom two rows clearly show that the proposed DSP block achieves about 9 times reduction in parameters compared with the one based on regular convolutions.

\begin{figure}[!t]
\centering
\includegraphics[width=3.3in]{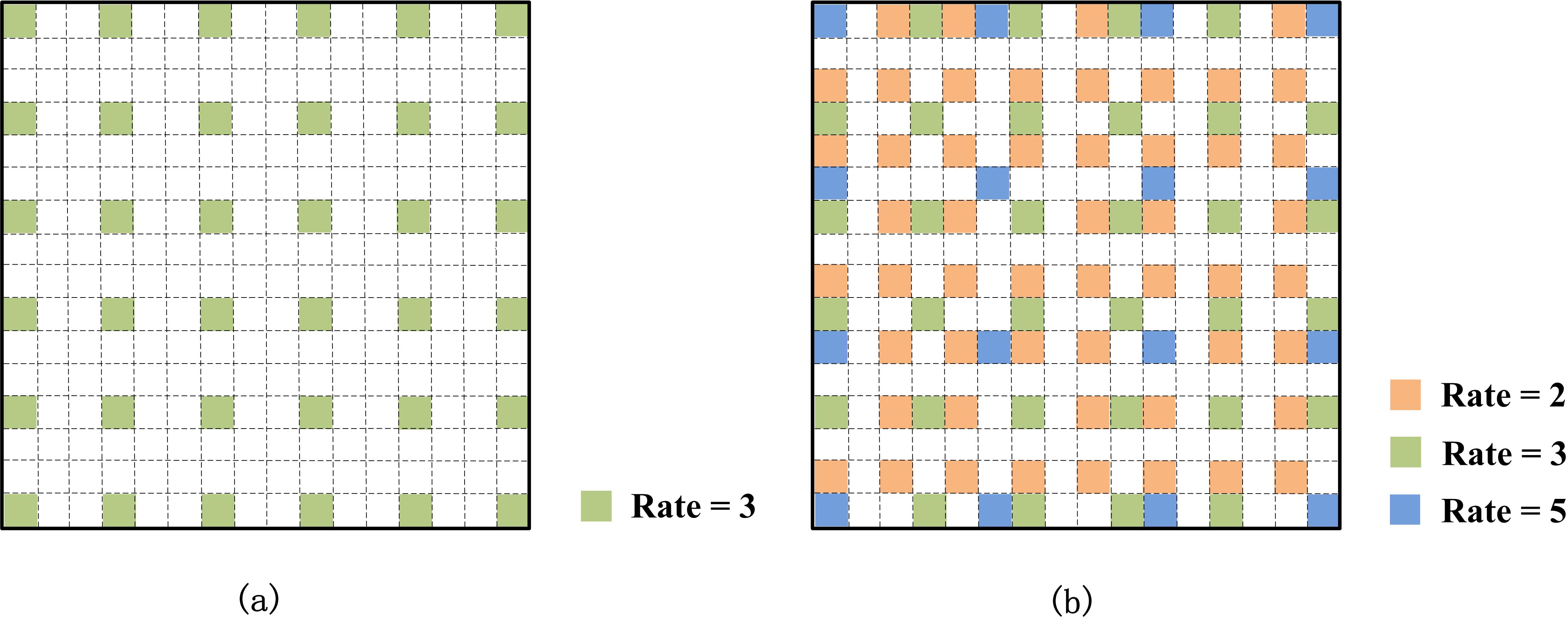}
% where an .eps filename suffix will be assumed under latex,
% and a .pdf suffix will be assumed for pdflatex; or what has been declared
% via \DeclareGraphicsExtensions.
\caption{Dilation sampling comparison. (a) Sparse Feature Sampling. (b) Dense Feature Sampling.}
\label{5}
\end{figure}

$\textbf{Denser dilation sampling.}$ Thereafter, all neurons in each branch share the same field of view at a single scale, which is hard to deal with scale variation cases for object parsing. Different from the previous sparse feature sampling strategy \cite{5}, \cite{13}, we incorporate contextual information at multiple dense scales, because pixels near the target usually contain more useful semantic information. To gather the local context information as much as possible, we successively replace original dilation rates in 3 $\times$ 3 depth-wise convolutions with a group of coprime dilation rates \textit{D} = $\{\textit{d$_{1}$, d$_{2}$, $\cdots$, d$_{n}$}\}$ (inspired by the prior work \cite{61}). As depicted in Fig. \ref{5}(b), a much denser feature sampling measure is provided to capture more relevant sub-regions \textit{L} = $\{\textit{l$_{1}^{H \times W \times (C_{o} \times r),d_{1}}$, l$_{2}^{H \times W \times (C_{o} \times r),d_{2}}$, $\cdots$, l$_{n}^{H \times W \times (C_{o} \times r),d_{n}}$}\}$ compared with the case in Fig. \ref{5}(a). In this way, the DSP block could enrich feature representations without any extra parameters. Although channel pruning operation dramatically decreases the computation cost, the difficulty in too much information loss becomes a major issue. To alleviate the performance degradation, we reasonably set \textit{r} to 0.2 and \textit{n} to 4 by performing ablation studies. The above process can be formulated as:
\begin{equation} \label{eq:4}
I^{'}{^{H \times W \times (C_{o} \times r)}} = \delta(w^{1 \times 1} \times I^{H \times W \times C_{i}} +b),
\end{equation}
\begin{equation} \label{eq:5}
\begin{split}
l_{k}^{H \times W \times (C_{o} \times r), d_{k}} = DSC_{d_{k}}(I^{'}{^{H \times W \times (C_{o} \times r)}}).
\end{split}
\end{equation}
where \textit{l}$_{k}^{H \times W \times (C_{o} \times r), d_{k}}$ refers to the \textit{k}-th path feature maps in the DSP block, \textit{k} $\in$ \{1, 2, $\cdots$, \textit{n}\}. \textit{DSC$_{d_{k}}$} ($\cdot$) denotes the Depthwise Separable Convolution with dilation rate \textit{d$_{k}$}.

$\textbf{Local and global context extraction.}$ The above design of DSP block is able to perceive patches or pixels locally, but falls short of a global view, especially for large objects and confusion categories. To remedy this defect, we add a path with small computation overheads, and then generate the global vectors \textit{G$^{1 \times 1 \times (C_{o} \times r)}$} from feature maps \textit{I$^{'}$$^{H \times W \times (C_{o} \times r)}$} by adopting a global average pooling (GAP) layer, similar to \cite{13}. Specifically, GAP performs down-sampling operation by computing the mean of the height \textit{H} and width \textit{W} dimensions of the input. Subsequently, the up-sampling layer utilizes the weighted average of two translated pixel values for each output pixel value of \textit{G$^{'}$$^{H \times W \times (C_{o} \times r)}$} by bilinear interpolation. After that, the features are integrated from all parallel paths. Please note that directly integrating them stemmed from a group convolution layer may weaken feature representations \cite{28}. Consequently, the channel shuffling operation is applied to ease cross-group information \textit{O$^{H \times W \times C_{o}}$}, as formulated below:
\begin{equation} \label{eq:8}
G{^{1 \times 1 \times (C_{o} \times r)}} = GAP(I^{'}{^{H \times W \times (C_{o} \times r)}}),
\end{equation}
\begin{equation} \label{eq:9}
G^{'}{^{H \times W \times (C_{o} \times r)}} = U(G{^{1 \times 1 \times (C_{o} \times r)}}),
\end{equation}
\begin{equation} \label{eq:9}
\begin{split}
O^{H \times W \times C_{o}} = S(C[L{^{H \times W \times (C_{o} \times r) \times n}}, G^{'}{^{H \times W \times (C_{o} \times r)}}]),
\end{split}
\end{equation}
where \textit{GAP}($\cdot$) denotes a global average pooling layer, \textit{U}($\cdot$) represents the bilinear interpolation operation. \textit{C}[$\cdots$] means the channel concatenation, and \textit{S}($\cdot$) corresponds to the channel shuffling operation.

$\textbf{Short-skip residual learning.}$ To promote the flows of contextual information throughout the network, we adopt a shortcut connection that strengthens gradient back-propagation in the DSP block, inspired by the principle of ResNet \cite{62}. Formally, the combination of input \textit{I$^{H \times W \times C_{i}}$} and non-linear transformation output \textit{O$^{H \times W \times C_{o}}$} is described as:
\begin{equation} \label{eq:10}
O^{'}{^{H \times W \times C_{o}}} = I^{H \times W \times C_{i}} \oplus O^{H \times W \times C_{o}} \ s.t. \ C_{i} \equiv C_{o}.
\end{equation}

\subsubsection{Going Deeper with Global View}
In contrast to previous multi-scale networks \cite{4}-\cite{6} that calculate large number of feature maps in a parallel-branch module, we apply three DSP blocks in cascade rather than in parallel to save computation resources and adopt image-level features in MCIM to enlarge the field of view, as illustrated in Fig. \ref{3}.

$\textbf{Multi-scale object variations.}$ We follow the strategy of pixel sampling rates in ASPP but keep dilation rates denser in the DSP block. Fig. \ref{3} presents that each DSP block has its own function for object parsing. Specifically, the first DSP block with a group of hybrid dilation rates \textit{D$_{s}$} = \{1, 2, 3, 5\} mainly focuses on small size objects (e.g., traffic signs). On the basis of the former, we employ the second DSP block with a group of hybrid dilation rates \textit{D$_{m}$} = \{7, 9, 11, 13\} to focus on medium size objects (e.g., cars). Further, the last DSP block with a group of hybrid dilation rates \textit{D$_{l}$} = \{17, 19, 21, 23\} is designed to focus on large size objects (e.g., buildings). It is worth mentioning that since high-level feature maps contain a limited number of channels and the resolution of feature maps is small, thus setting large dilation rate does not increase too much computation overheads. Finally, we successively combine the output of each DSP block by element-wise sum to jointly encode multi-level semantics for better recognition performance, similar to FCN \cite{29}.

$\textbf{Enlarging the field of view.}$ As formulated in Eq. (10) and Eq. (11), each convolution layer with dilation rate \textit{D} and kernel size \textit{K} in the DSP block could obtain the field of view \textit{R}. Therefore, stacking all DSP blocks together can theoretically obtain the largest field of view \textit{R$_{max}$}:
\begin{equation} \label{eq:11}
R = (D-1)\times(K-1)+K,
\end{equation}
\begin{equation} \label{eq:12}
R_{max} = R^{s}_{max} + R^{m}_{max} + R^{l}_{max}-2,
\end{equation}
for instance, the MCIM module will obtain the field of view with maximum size 83.

We argue that the size of effective field of view is practically smaller, since amounts of information (e.g., marginal areas of feature maps) is abandoned as the network goes deeper. To handle this problem, we add a GAP layer on the tail of the backbone to enlarge the field of view. The global information \textit{G$^{''}$${^{H \times W \times C}}$} is extracted from \textit{F$^{H \times W \times C_{i}}_{Stage_{7}}$}, which keeps the valid field of view large enough from a macroscopic aspect. Then we combine \textit{G$^{''}$${^{H \times W \times C}}$} with integrated semantics \textit{O}$^{''}$ = \{\textit{O}$^{'}$$^{H \times W \times C_{o}}_{1}$ $\oplus$ \textit{O}$^{'}$$^{H \times W \times C_{o}}_{2}$ $\oplus$ \textit{O}$^{'}$$^{H \times W \times C_{o}}_{3}$\} to generate the final output of MCIM. This process can be expressed as:
\begin{equation} \label{eq:13}
G^{''}{^{H \times W \times C}} = U(GAP(\delta(w^{1 \times 1} \times F^{H \times W \times C_{i}}_{Stage_{7}}+b))),
\end{equation}
%\begin{equation} \label{eq:14}
%O^{''}{^{H \times W \times C_{o}}} = O^{'}_{1}{^{H \times W \times C_{o}}} \oplus O^{'}_{2}{^{H \times W \times C_{o}}} \oplus O^{'}_{3}{^{H \times W \times C_{o}}}
%\end{equation}
\begin{equation} \label{eq:14}
Y{^{H \times W \times (C_{o} + C)}} = C[O^{''}{^{H \times W \times C_{o}}}, G^{''}{^{H \times W \times C}}].
\end{equation}

\section{Experimental Evaluation}
To evaluate the performance of CIFReNet, we detailedly present an experimental procedure as below.
\subsection{Training Protocol}
\subsubsection{Datasets}
$\textbf{Cityscapes Dataset.}$ The Cityscapes is a popular dataset for urban object parsing \cite{31}. It consists of 25,000 annotated 2,048 $\times$ 1,024 resolution images. The fine-annotated dataset contains 5000 images including 19 valid classes. There are 2,975 images for training, 500 images for validation, and 1,525 images for testing. Note that the ground truth of testing set is unavailable, we shall submit the results to an online test server. Due to the limited physical memory, we randomly sub-sample image resolution to 1,024 $\times$ 512 patches for training.

$\textbf{CamVid Dataset.}$ The CamVid is another urban object parsing dataset \cite{53} which contains 12 classes in total. It consists of 367 frames, 101 frames, and 223 frames for training, validation, and testing, respectively. The original frame resolution of CamVid is 960 $\times$ 720. Following the prior works \cite{8}, \cite{16}, we randomly sub-sample all images to 480 $\times$ 360 patches for training.

$\textbf{Helen Dataset.}$ The Helen is a widely-used dataset for facial object parsing \cite{78}, which consists of 2,330 face images with 11 labeled categories including face, nose, left eye, right eye, left eyebrow, right eyebrow, upper lip, inner mouth, lower lip, hair and background. Helen dataset is divided into 2,000 images, 230 images, and 100 images for training, validation and, testing. We randomly rescale the image resolution to 512 $\times$ 512 patches for training.

\subsubsection{Metrics}
$\textbf{Segmentation accuracy.}$ The Mean Intersection over Union (MIoU) is commonly adopted for semantic segmentation accuracy. Suppose \textit{n} is the number of classes, we can compute it as:
\begin{equation} \label{eq:5}
MIoU = \frac{1}{n}\sum_{i=1}^n\frac{p_{ii}}{\sum_{j=1}^np_{ij}+\sum_{j=1}^np_{ji}-p_{ii}},
\end{equation}
where \textit{p$_{ij}$} is the number of pixels that belong to \textit{i} predicted to class \textit{j}, \textit{p$_{ii}$} refers to the true positive. \textit{p$_{ij}$} and \textit{p$_{ji}$} refer to the false positive and false negative, respectively.

$\textbf{Execution speed.}$ The amount of forward pass time in millisecond (ms) that a network takes to process an image, which is generally measured by frames per second (FPS).

$\textbf{Network parameters.}$ The total number of weights and biases of each layer in the network.

$\textbf{Memory footprint.}$ The storage space that is required to store the network parameters.

$\textbf{Computational complexity.}$ The number of float-point operations (FLOPs) for measuring how quickly and effectively a CNN model works.

\subsubsection{Training Details}
We implement all experiments in Pytorch with NVIDIA 1080Ti GPU cards. All ablation results are evaluated on Cityscapes validation set. To avoid the over-fitting, we randomly scale images from 0.5 to 1.5 ratio and left-right flip them for all datasets. Besides, we employ the mean subtraction and add a random rotation operation from -3 to 3 degrees during training process. Following the prior protocol \cite{4}, we set initial learning rate to 0.005 and employ ``Poly" learning rate policy by $1-(\frac{iter}{max\_iter})^{power}$  with a power 0.9. In the end-to-end learning, the network is trained by using Stochastic Gradient Descent (SGD) optimization algorithm, of which the momentum is 0.9 and weight decay is 5e-4. In addition, the pixel-wise cross-entropy error is employed as our loss function.

\subsection{Modified MobileNet V2}
\subsubsection{Ablation for Backbone Choices}
Designing a novel backbone from scratch often requires expensive training resources, so we compare the pretrained MobileNet V2 with other pretrained backbone networks that are widely adopted in current segmentation models. For fairness, all experiments are carried out on a FCN-based model. As shown in Table \ref{IV}, FCN-VGGNet16 is obviously not comparable to FCN-MobileNet V2 in both accuracy and efficiency due to the limited performance and heavy structure of itself. Moreover, though FCN-ResNets achieve similar or better accuracy compared with FCN-MobileNet V2, the former generally contain enormous parameters and require more memory resources. Additionally, we observe that FCN-DenseNet121 is a pretty good choice, which yields a solid MIoU score of 68.41\% while the requirement of resources is far less than FCN-ResNets. Further, FCN-DenseNet121 outperforms FCN-MobileNet V2 in terms of accuracy by 0.75\% MIoU, but incurs about 4 times parameters and storage complexity increase. To sum up, all above comparisons demonstrate that employing MobileNet V2 as feature extractor could more efficiently bring the great benefit for object parsing.

\begin{table}[!t]\caption{Ablation studies on backbone choices. $\sharp$Params: The number of parameters. $\sharp$NS: Network Size. }
\centering
  \begin{tabular}{c|c|c|c}
  \toprule[1pt]
  Method&$\sharp$Params(M)&$\sharp$NS(MB)&MIoU(\%)\\
  \midrule[0.5pt]
  FCN-VGGNet16	&134.35	&524.82& 62.29\\
  FCN-ResNet101& 42.67&	167.18& 70.70\\
  FCN-ResNet50	&23.68&	92.75&	67.78\\
  FCN-ResNet18	&11.30&	44.19&	66.82\\
  FCN-DenseNet121&	7.09&	28.11&	68.41\\
  \midrule[0.5pt]
  FCN-MobileNet V2&	1.93&	7.68&	67.66\\
  \bottomrule[1pt]
  \end{tabular}
\label{IV}
\end{table}
\begin{table}[!t]\caption{Ablation studies on initial settings. OS: Output Stride.}
\centering
  \begin{tabular}{c|c|c|c}
  \toprule[1pt]
  OS = 8&OS = 16&OS = 32&MIoU(\%)\\
  \midrule[0.5pt]
  $\surd$& & & 59.41\\
   & $\surd$& &61.39\\
    & & $\surd$& 60.75 \\
  \bottomrule[1pt]
  \end{tabular}
\label{V}
\end{table}
\begin{figure}[!t]
\centering
\includegraphics[width=3.5in]{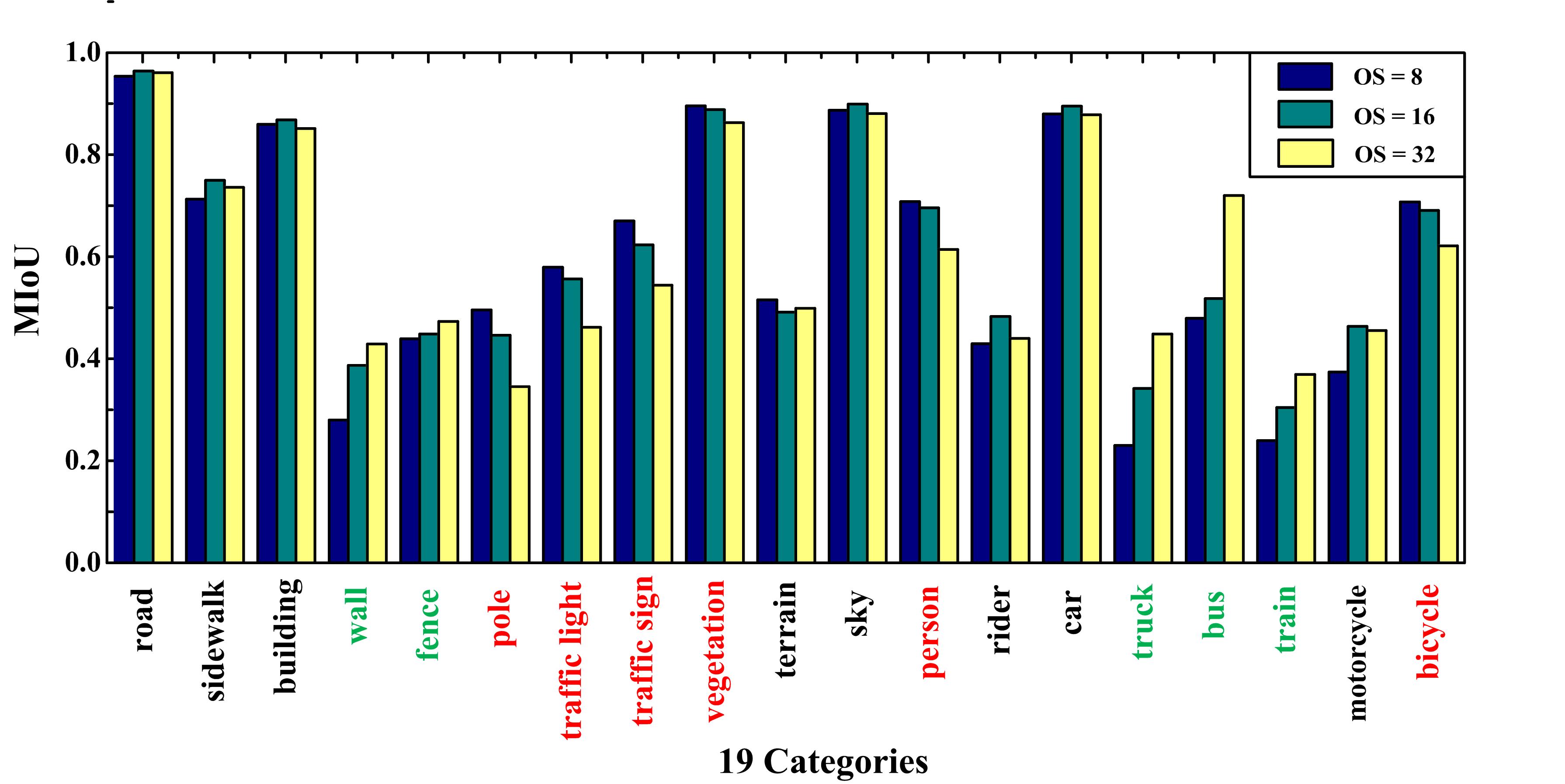}
% where an .eps filename suffix will be assumed under latex,
% and a .pdf suffix will be assumed for pdflatex; or what has been declared
% via \DeclareGraphicsExtensions.
\caption{The performance of 19 categories with different OS settings on Cityscapes validation set. Note that the categories in Red imply that setting smaller OS could boost the recognition of relatively small objects, and the categories in Green imply that setting larger OS could boost the recognition of relatively obvious ones.}
\label{6}
\end{figure}
\subsubsection{Ablation for Output Stride}
In this part, we conduct several experiments to explore an appropriate output stride (OS) value. Since processing high-resolution images (3 $\times$ 2,048 $\times$ 1,024) limits GPU resources, we do not consider further denser feature maps (e.g., OS $\leqslant$ 4). As seen in Table \ref{V} and Fig. \ref{6}, employing OS = 16 or 32 obtains a bit higher accuracy than the model with OS = 8, especially for obvious objects marked with green color. It reflects that successive sub-sampling layers could bring sufficient field of view and make features more discriminative. Unfortunately, the performance of larger OS in small objects marked with red color are unsatisfied, resulting from amounts of information loss when setting OS to 16 or 32. Therefore, the value of OS is set to 8, so as not to make an irreversible affect for spatial details recovery during model inference.
\begin{table}[!t]\caption{ Ablation studies on initial settings. HDRS: Hybrid Dilation Rates Strategy. UL: Up-sampling Layer. DA: Data Augmentation. FLOPs are estimated on a 3 $\times$ 640 $\times$ 360 input.}
\centering
  \begin{tabular}{c|c|c|c|c}
  \toprule[1pt]
  HDRS&UL&DA&FLOPs(G)&MIoU(\%)\\
  \midrule[0.5pt]
  \{1, 1, 1, 1\}&& &6.89&59.41\\
  \{2, 2, 4, 4\}&& &6.92 (0.03$\blacktriangle$)&66.15 (6.74$\blacktriangle$)\\
  \{1, 2, 3, 5\}&& &6.91 (0.02$\blacktriangle$)&66.17 (6.76$\blacktriangle$)\\
  \{2, 3, 5, 7\}&& &6.94 (0.05$\blacktriangle$)&67.36 (7.95$\blacktriangle$)\\
  \{3, 5, 7, 11\}&& &6.97 (0.08$\blacktriangle$)&69.41 (10.0$\blacktriangle$)\\
  \midrule[0.5pt]
  \multirow{2}{*}{\{2, 3, 5, 7\}}& &$\surd$&6.94 (0.05$\blacktriangle$)&67.57 (8.16$\blacktriangle$)\\
  &$\surd$&$\surd$&6.95 (0.06$\blacktriangle$)&67.84 (8.43$\blacktriangle$)\\
  \bottomrule[1pt]
  \end{tabular}
\label{VI}
\end{table}
\begin{figure}[!t]
\centering
\includegraphics[width=3.3in]{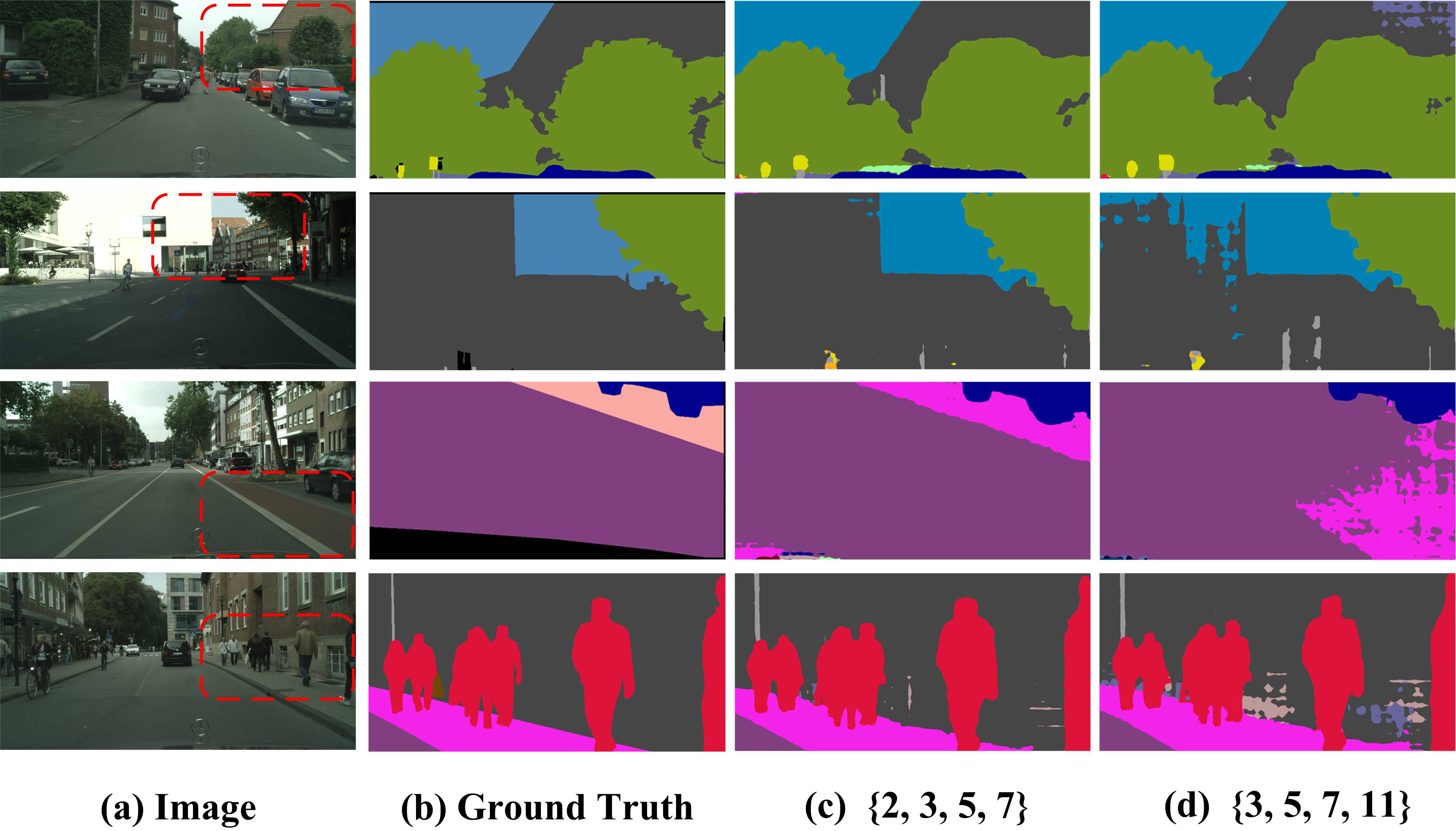}
% where an .eps filename suffix will be assumed under latex,
% and a .pdf suffix will be assumed for pdflatex; or what has been declared
% via \DeclareGraphicsExtensions.
\caption{Visualization results of ``Gridding Effect". From left to right: Image, Ground Truth, the group of hybrid dilation rates \{2, 3, 5, 7\} and \{3, 5, 7, 11\} added into the backbone (best viewed in color).}
\label{7}
\end{figure}

\begin{table*}[!t]\caption{Performance comparisons with various long-skip modules. \textit{L$_{n}$}: Low-frequency information outputted from \textit{n}-th stage, \textit{n} $\in$ (1, 2, 3). \textit{H$_{n}$}: High-frequency information outputted from \textit{n}-th stage, \textit{n} $\in$ (6, 7). CA: Channel Attention. FPT: Forward Pass Time. $\sharp$Params: The number of parameters. Note that FPT and FLOPs are estimated on a 3 $\times$ 640 $\times$ 360 input.}
\centering
  \begin{tabular}{c|c|c|c|c|c|c|c|c|c|c|c}
  \toprule[1pt]
  Model&\textit{L$_{1}$}&\textit{L$_{2}$}&\textit{L$_{3}$}& \textit{H$_{6}$}& \textit{H$_{7}$} &CA&Sum& FPT(ms)&$\sharp$Params(M) & FLOPs(G) &MIoU(\%)\\
  \midrule[0.5pt]
  Baseline& & & & & & & &13.52&1.82&6.95& 67.84\\
  \textit{a}&$\surd$& & &$\surd$& &$\surd$&&14.83 (1.31$\blacktriangle$)&1.83 (0.01$\blacktriangle$)&6.96 (0.01$\blacktriangle$)& 68.43 (0.59$\blacktriangle$)\\
  \textit{b}& & $\surd$& &$\surd$& &$\surd$& &14.20 (0.68$\blacktriangle$)&1.83 (0.01$\blacktriangle$)&6.97 (0.02$\blacktriangle$)& 68.32 (0.48$\blacktriangle$)\\
  \textit{c}& & & $\surd$&$\surd$& &$\surd$& &14.11 (0.59$\blacktriangle$)&1.83 (0.01$\blacktriangle$)&6.97 (0.02$\blacktriangle$)& 68.68 (0.84$\blacktriangle$)\\
  \textit{d}& & & $\surd$& & $\surd$&$\surd$& &14.51 (0.99$\blacktriangle$)&1.85 (0.03$\blacktriangle$)&6.99 (0.04$\blacktriangle$)& 68.82 (0.98$\blacktriangle$)\\
  \textit{e}& & & $\surd$ &$\surd$ & & & $\surd$ &14.04 (0.52$\blacktriangle$)&1.83 (0.01$\blacktriangle$)&6.97 (0.02$\blacktriangle$)& 68.02 (0.18$\blacktriangle$)\\
  \bottomrule[1pt]
  \end{tabular}
\label{VII}
\end{table*}
\subsubsection{Ablation for Hybrid Dilation Rates Strategy and Baseline}
To evaluate the performance of dilation strategy in the backbone, we compare several proposals with the vanilla case where HDRS = \{1, 1, 1, 1\}. Note that the abbreviation HDRS refers to the hybrid dilation rates strategy applied in our backbone from Stage$_{4}$ to Stage$_{7}$. As summarized in Table. \ref{VI}, we find out that \textit{$\textbf{a}$}) Compared with the original one, replacing the last four sub-sampling layers with dilation convolutions generally yields about 6\%-10\% MIoU boost, which illustrates the effectiveness of our strategy for backbone design. \textit{$\textbf{b}$}) Adopting HDRS = \{2, 2, 4, 4\} indeed benefits the accuracy performance, but it is not the best choice. Due to same or geometric dilation ratios, the higher layer mostly samples information in the same region as the former one, thus making the field of view restricted. \textit{$\textbf{c}$}) The model obtains better segmentation results as the group of dilation rates becomes larger. However, the filters become too sparse to cover any relevant information, which requires more computation resources and leads to the aggravation of ``Gridding Issue". The red box of Fig. \ref{7} intuitively presents the comparison between HDRS = \{2, 3, 5, 7\} and HDRS = \{3, 5, 7, 11\} about this issue. For further performance improvement, we suggest that employing HDRS = \{2, 3, 5, 7\} to ensure both accuracy and efficiency. Additionally, instead of directly up-sampling the final feature maps by a simple convolution layer, we elaborately design an up-sampling layer as a transition operation for better information recovery. Further results show that the baseline model with up-sampling layer and data argumentation yields higher performance up to 67.84\% MIoU, while the extra computation cost is negligible.
\begin{figure}[!t]
\centering
\includegraphics[width=3.3in]{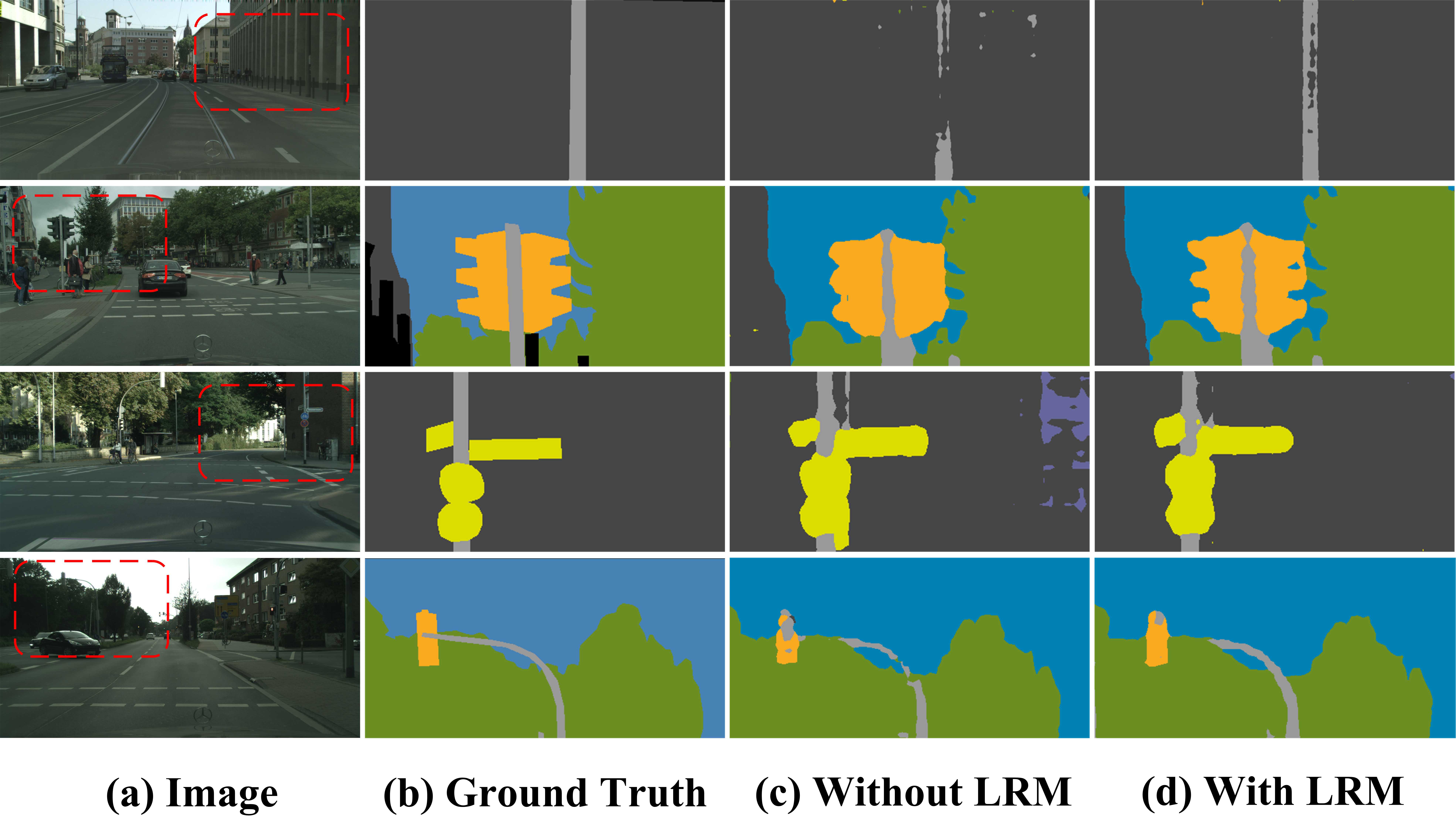}
% where an .eps filename suffix will be assumed under latex,
% and a .pdf suffix will be assumed for pdflatex; or what has been declared
% via \DeclareGraphicsExtensions.
\caption{Visualization results of LRM on Cityscapes validation set. From left to right: Image,
Ground Truth, Without LRM, With LRM (best viewed in color).}
\label{8}
\end{figure}

\subsection{LRM and MCIM}
Here we evaluate the performance of Long-skip Refinement Module and Multi-scale Context Integration Module.

\subsubsection{Ablation for Long-skip Refinement Module}
\begin{figure}[!t]
\centering
\includegraphics[width=3.5in]{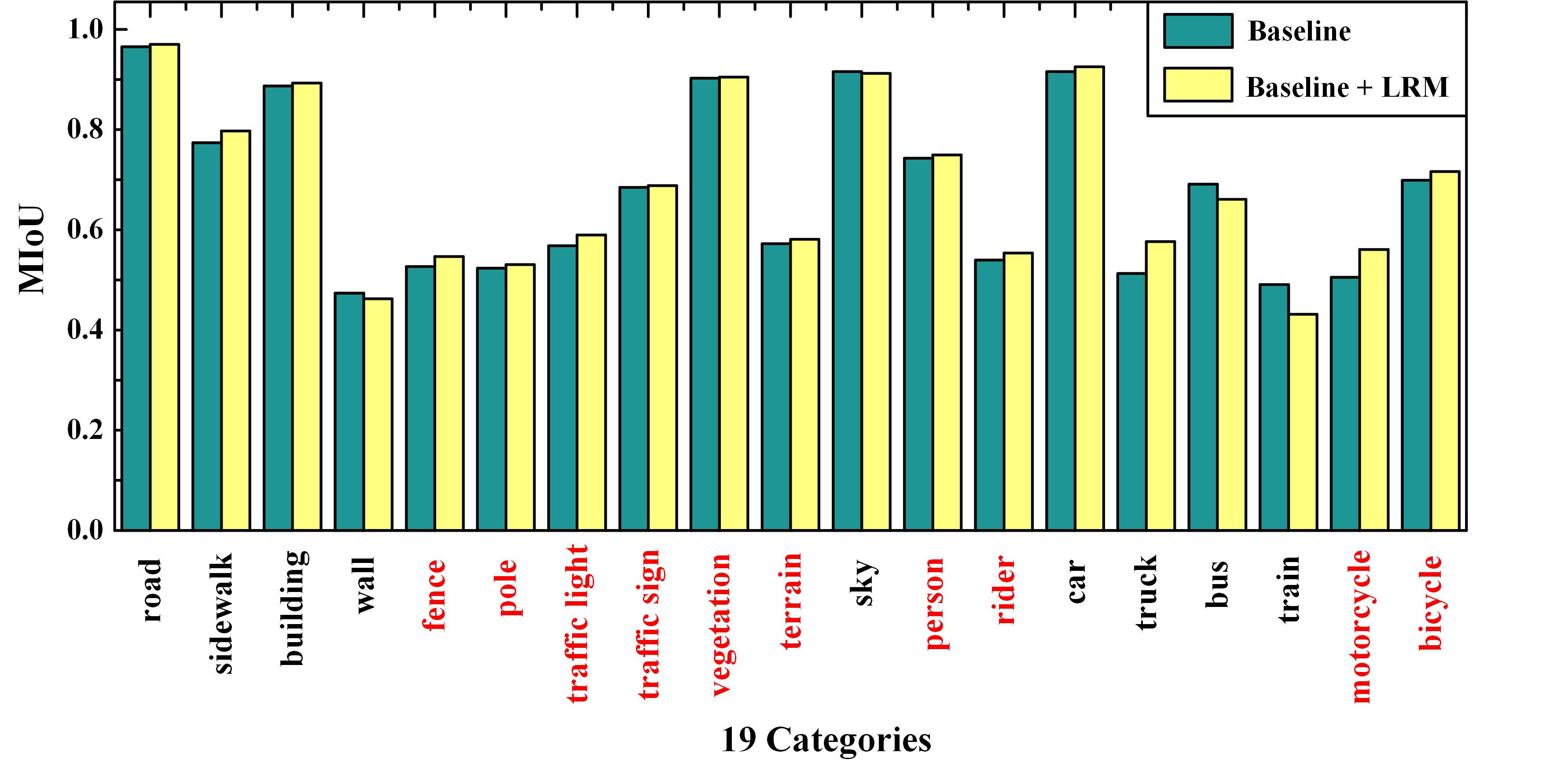}
% where an .eps filename suffix will be assumed under latex,
% and a .pdf suffix will be assumed for pdflatex; or what has been declared
% via \DeclareGraphicsExtensions.
\caption{The performance of Baseline and Baseline+LRM with respect to 19 categories on Cityscapes validation set. Note that the categories in Red refer to relatively small objects.}
\label{9}
\end{figure}

As shown in Table \ref{VII}, we carry out several ablation studies to reveal the effectiveness of LRM. Note that the check marks refer to the two stages united by a long-skip connection. We firstly fix the semantic feature layer (e.g., Stage$_{6}$) and then vary the shallow feature layer (e.g., Stage$_{1}$ to Stage$_{3}$), so as to evaluate the quality of various low-frequency features during the attention refinement process. As seen, Table \ref{VII} has demonstrated that methods based on long-ship learning strategy all outperform the baseline method. Nevertheless, the performance of these methods are not similar due to their distinctive structures. Specifically, we observe that module \textit{c} achieves comparable or slightly better performance in terms of both accuracy and efficiency than module \textit{a} and module \textit{b}. We attribute the superiority of module \textit{c} as two aspects: \textit{$\textbf{a}$}) Different from module \textit{a} and module \textit{b} where shallow layers contain large amounts of noise in feature maps, module \textit{c} could generate purer spatial information after deeper convolution layers, which is more effective for the attention refinement process and the later feature fusion. \textit{$\textbf{b}$}) Module \textit{c} achieves a faster inference and avoids some needless computation cost without any extra down-sampling operation on high resolution feature maps. Moreover, although module \textit{d} outperforms module \textit{c} in terms of accuracy, we also observe that the former brings more than 2 times computation cost increase compared with the latter one, which is inconsistent with our purpose of a lightweight design. In addition, the accuracy of module \textit{e} where a straightforward fusion of low-level and high-level features is implemented by element-wise sum yields only 0.18\% MIoU boost. This indicates that such an manner may limitedly improve recognition performance, because an excessive semantic gap between low-frequency and high-frequency features helps little for spatial information learning. Based on all above observations, we regard module \textit{c} as the proposed LRM, which achieves 0.84\% MIoU improvement with 0.01 M extra parameters and 0.02 GFLOPs increase compared with the baseline method. Also, these results well demonstrates the effectiveness of LRM in terms of both accuracy and efficiency. The visual results comparison between the baseline method and the one with LRM is provided in Fig. \ref{8} and Fig. \ref{9}, which also shows that some small objects marked with red color (e.g., poles, traffic signs) could benefit from the proposed LRM.

\begin{table*}[!t]\caption{
Comparisons of cascaded DSP blocks with different settings when adopting modified MobileNet V2 as backbone. \textit{n}: The number of paths in DSP. \textit{r}: Channel reduction ratio. \textit{D$_{k}$} represents sampling scales of DSP$_{\textit{k}}$ block, \textit{k} $\in$ (\textit{s}, \textit{m}, \textit{l}). GAP: Global Average Pooling. $\sharp$Params: The number of parameters. Note that FLOPs are estimated on a 320 $\times$ 256 $\times$ 128 input.}
\centering
  \begin{tabular}{c|c|c|c|c|c|c|c|c}
  \toprule[1pt]
  \textit{n}&\textit{r}&\textit{D$_{s}$}&\textit{D$_{m}$}&\textit{D$_{l}$}&GAP&$\sharp$Params(M)&FLOPs(G)&MIoU(\%)\\
  \midrule[0.5pt]
  2&1/2&\{6$_{1}$, 6$_{2}$, $\cdots$, 6$_{n}$\}&\{12$_{1}$, 12$_{2}$, $\cdots$, 12$_{n}$\}&\{18$_{1}$, 18$_{2}$, $\cdots$, 18$_{n}$\}& &0.21&6.73&70.48 \\
  4&1/4&\{6$_{1}$, 6$_{2}$, $\cdots$, 6$_{n}$\}&\{12$_{1}$, 12$_{2}$, $\cdots$, 12$_{n}$\}&\{18$_{1}$, 18$_{2}$, $\cdots$, 18$_{n}$\}& &0.11&3.58&70.89\\
  8&1/8&\{6$_{1}$, 6$_{2}$, $\cdots$, 6$_{n}$\}&\{12$_{1}$, 12$_{2}$, $\cdots$, 12$_{n}$\}&\{18$_{1}$, 18$_{2}$, $\cdots$, 18$_{n}$\}&&0.06&2.00&70.53\\
  16&1/16&\{6$_{1}$, 6$_{2}$, $\cdots$, 6$_{n}$\}&\{12$_{1}$, 12$_{2}$, $\cdots$, 12$_{n}$\}&\{18$_{1}$, 18$_{2}$, $\cdots$, 18$_{n}$\}& &0.04&1.21&70.23\\
  4&1/4&\{1, 2, 3, 5\}&\{7, 9, 11, 13\}&\{17, 19, 21, 23\}& &0.11&3.57&71.58\\
  4&1/(4+1)&\{1, 2, 3, 5\}&\{7, 9, 11, 13\}&\{17, 19, 21, 23\}&$\surd$&0.09&2.47&71.90\\
  \bottomrule[1pt]
  \end{tabular}
\label{VIII}
\end{table*}
\subsubsection{Ablation for Dense Semantic Pyramid Block}
To investigate the effect on different settings of \textit{n} and \textit{r} for model performance, we conduct some experiments under the same condition. Based on the results in Table \ref{VIII}, we draw several conclusions. On the one hand, with the number of path \textit{n} increases, network parameters and computational complexity gradually decrease and eventually suffer from the accuracy degradation from 70.89\% MIoU to 70.23\% MIoU. These results illustrate that excessive channel reduction could damage the accuracy performance due to large amounts of information loss. On the other hand, adopting diverse sampling rates to capture multi-scale local context information near the target in a dense manner yields 0.69\% MIoU boost, which proves that the proposed dense feature sampling strategy delivers a stronger capacity of feature representations in DSP block. Furthermore, the proposed DSP block where both local and global context information are jointly encoded yields about 0.32\% MIoU improvement compared with the one without image-level features. According to these observations, the optimal setting for \textit{n} is 4 and \textit{r} is 0.2, and the proposed DSP block achieves high accuracy of 71.90\% MIoU with only 0.09 M parameters and 2.47 GFLOPs increase.

\begin{table}[!t]\caption{Ablation studies on MCIMs. MCIM(C): DSP blocks in cascade. MCIM(P): DSP blocks in parallel.}
\centering
  \begin{tabular}{c|c|c|c|c c}
  \toprule[1pt]
  \multirow{2}{*}{Model}&\multicolumn{2}{|c|}{MCIM(C)}&\multirow{2}{*}{MCIM(P)}&\multirow{2}{*}{MIoU(\%)}\\ \cline{2-3}
   &DSP(C)&GAP & & \\
  \midrule[0.5pt]
  \multirow{4}{*}{Baseline}& & & & 67.84\\
  &$\surd$& & &71.90 (4.06$\blacktriangle$)\\
  & & & $\surd$&72.42 (4.58$\blacktriangle$)\\
  & $\surd$& $\surd$& &72.14 (4.30$\blacktriangle$)\\
  \bottomrule[1pt]
  \end{tabular}
\label{IX}
\end{table}
\begin{table}[!t]\caption{
Performance comparisons with other state-of-the-art multi-scale modules. FPT: Forward Pass Time. $\sharp$Params: The number of parameters. Note that FPT and FLOPs are estimated on a 320 $\times$ 256 $\times$ 128 input.}
\centering
  \begin{tabular}{c|c|c|c|c}
  \toprule[1pt]
  Model&FPT(ms)& $\sharp$Params(M)& FLOPs(G) &MIoU(\%)\\
  \midrule[0.5pt]
  Vortex \cite{63}&111.33&5.54&230.29&72.24\\
  ASPP \cite{13}&67.92&4.52&174.86&70.87\\
  FPA \cite{6}&28.49&7.48&43.30&71.92\\
  PPM \cite{4}&19.28&1.28&25.21&71.00\\
  \midrule[0.5pt]
  MCIM(P)&27.79&0.48&15.06&72.42\\
  MCIM(C)&28.58&0.12&2.48&72.14\\
  \bottomrule[1pt]
  \end{tabular}
\label{X}
\end{table}
\subsubsection{Ablation for Multi-scale Context Integration Module} As shown in Table \ref{IX}, the MCIM(C) denotes that the DSP block are designed in cascade with global information and the MCIM(P) denotes that the DSP blocks are designed in parallel without global information. The model based on MCIM(P) outperforms the model based on DSP(C) module by 0.52\% MIoU under the same training conditions, which indicates that the cascaded structure suffers from the feature degradation as the network goes deeper. To overcome this problem, we incorporate image-level features into the DSP(C) module to enhance the model learning capacity. It is clear that the model based on MCIM(C) achieves 72.14\% MIoU eventually. In the following, the overall performance of both MCIM(C) and MCIM(P) will be discussed.
\begin{table}[!t]\caption{Ablation studies on LRM and MCIM(C). LRM: Long-skip Refinement Module. MCIM(C): Multi-scale Context Integration Module.}
\centering
  \begin{tabular}{c|c|c|c}
  \toprule[1pt]
  Baseline&LRM&	MCIM(C)&MIoU(\%)\\
  \midrule[0.5pt]
  $\surd$& & &67.84\\
  $\surd$&$\surd$& &68.68 (0.84$\blacktriangle$)\\
  $\surd$& &$\surd$&72.14 (4.30$\blacktriangle$)\\
  $\surd$&$\surd$&$\surd$&72.95 (5.11$\blacktriangle$)\\
  \bottomrule[1pt]
  \end{tabular}
\label{XI}
\end{table}
\begin{figure}[!t]
\centering
\includegraphics[width=3.3in]{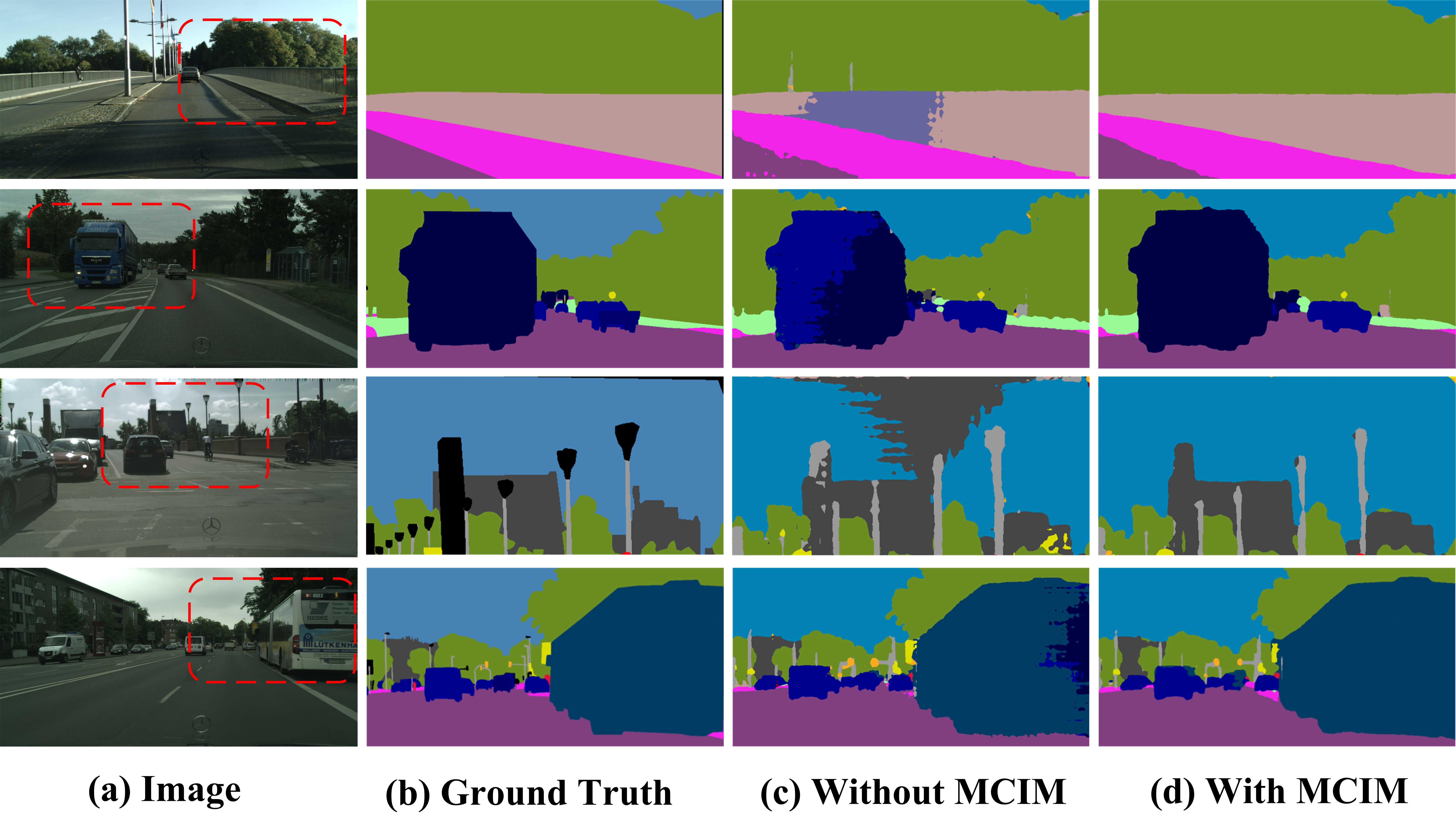}
% where an .eps filename suffix will be assumed under latex,
% and a .pdf suffix will be assumed for pdflatex; or what has been declared
% via \DeclareGraphicsExtensions.
\caption{Visualization results of MCIM on Cityscapes validation set. From left to right: Image,
Ground Truth, Without MCIM, With MCIM (best viewed in color).}
\label{10}
\end{figure}

\begin{table*}[!t]\caption{Performance comparisons on Cityscapes test set.}
\centering
  \begin{tabular}{c|c|c|c|c|c|c}
  \toprule[1pt]
  Method&Backbone&Resolution&ES(FPS)&FLOPs(G)&$\sharp$Params(M)&MIoU(\%)\\
  \midrule[0.5pt]
  SegNet \cite{8}&/&640 $\times$ 360&14.6&286.0&29.5&56.1\\
  ENet \cite{16}&/&640 $\times$ 360&135.4&3.8&0.4&58.3\\
  Skip-ShuffleNet$^{\dagger}$ \cite{34}&ShuffleNet V1&1024 $\times$ 512&-&6.2&1.0&58.3\\
  SQNet$^{\dagger}$\cite{14}&/&640 $\times$ 360&11.6&-&-&59.8\\
  Skip-MobileNet$^{\dagger}$ \cite{34}&MobileNet V1&1024 $\times$ 512&45.0&15.4&3.4&62.4\\
  ERFNet$^{\dagger}$ \cite{23}&/&1024 $\times$ 512&41.7&53.5&2.1&68.0\\
  ICNet \cite{24}&/&1024 $\times$ 512&46.3&-&26.5&69.5\\
  GUNet$^{\dagger}$ \cite{18}&/&1024 $\times$ 512&33.3&-&-&70.4\\
  LW-RefineNet101$^{\dagger}$ \cite{22}&ResNet101&512 $\times$ 512&55.0&52.0&46.0&72.1\\
  LinkNet$^{\dagger}$ \cite{10}&ResNet18&640 $\times$ 360&65.8&21.2&11.5&72.6$^{\ast}$\\
  L-DenseNet121$^{\dagger}$ \cite{66}&ResNet18&1024 $\times$ 448&31.0&9.8&-&72.8$^{\ast}$\\
  \midrule[0.5pt]
  DSPNet$^{\dagger}$ \cite{68}&ResNet50&1024 $\times$ 512&14.0&-&144.4&64.9$^{\ast}$\\
  FCN8s$^{\dagger}$\cite{29}&VGGNet16&1024 $\times$ 512&2.0&136.2&134.5&65.3\\
  Dilation10$^{\dagger}$\cite{69}&VGGNet16&1024 $\times$ 512&0.3&-&140.8&67.1\\
  DRN-C-26 \cite{70}&ResNet50&1024 $\times$ 512&-&355.2&20.6&68.0$^{\ast}$\\
  LRR-4x \cite{71}&VGGNet16&1024 $\times$ 512&-&-&-&69.7\\
  DeepLab V2 \cite{5}&ResNet101&1024 $\times$ 512&0.3&457.8&44.0&70.4\\
  DLC \cite{72}&IRNet&1024 $\times$ 512&-&26.5&-&71.1\\
  FRRNet \cite{73}&/&1024 $\times$ 512&0.3&235.0&-&71.8\\
  RefineNet \cite{9}&ResNet101&1024 $\times$ 512&0.9&118.1&-&73.6\\
  DenseASPP121 \cite{46}&DenseNet121&1024 $\times$ 512&-&155.8&28.6&76.2\\
  PSPNet \cite{4}&ResNet101&713 $\times$ 713&0.8&412.2&250.8&78.4\\
  \midrule[0.5pt]
  \multirow{5}{*}{CIFReNet$^{\dagger}$(Ours)}&\multirow{5}{*}{MobileNet V2}&640 $\times$ 360&62.5&7.3 &\multirow{5}{*}{1.9}&\multirow{5}{*}{72.9$^{\ast}$/70.9}\\
  & &512 $\times$ 512&55.6&8.3& & \\
  & &713 $\times$ 713&33.3&16.3& &  \\
  & &1024 $\times$ 448&38.5&14.4&& \\
  & &1024 $\times$ 512&34.5&16.5&& \\
  \bottomrule[1pt]
  \end{tabular}
\label{XII}
\end{table*}

\subsubsection{Comparison with Similar Multi-scale Modules}In this subsection, we investigate several powerful yet computationally expensive multi-scale modules and compare them with the proposed MCIMs. The last two rows in Table \ref{X} reveals that MCIMs drastically reduce the number of parameters and simplify the computational complexity while maintaining the competitive accuracy. Specifically, MCIMs achieve speed-faster performance with about 46 times decrease in parameters and about 93 times reduction in FLOPs compared with Vortex module and ASPP module. In the case of FPA module and PPM module, both of them deliver similar accuracy to MCIMs, while suffering from heavier overheads (at least 11 times parameters and 10 times computational complexity increase). These results indicate that current multi-scale modules face the challenge of large amounts of time-consuming operations and invalid computation. Furthermore, it is obvious that the model based on MCIM(P) obtains a bit better accuracy than the model based on MCIM(C), because the former establishes a wider multi-scale module which is helpful to improve the network learning capacity (as demonstrated in \cite{64}). Nevertheless, the ``deep and thin" design of MCIM(C) achieves 4 times parameters decrease and 6 times FLOPs reduction compared with MCIM(P). These comparisons illustrate that the MCIM(C) can more efficiently encode multiple context information to generate accurate results, and MCIM(P) is regarded as a sub-optimal choice. Fig. \ref{10} presents some visual examples where some confusion categories exist. Obviously, some analogous objects (e.g., walls and buildings, trucks and buses) are correctly identified by the proposed MCIM.
\begin{figure}[!t]
\centering
\includegraphics[width=3.5in]{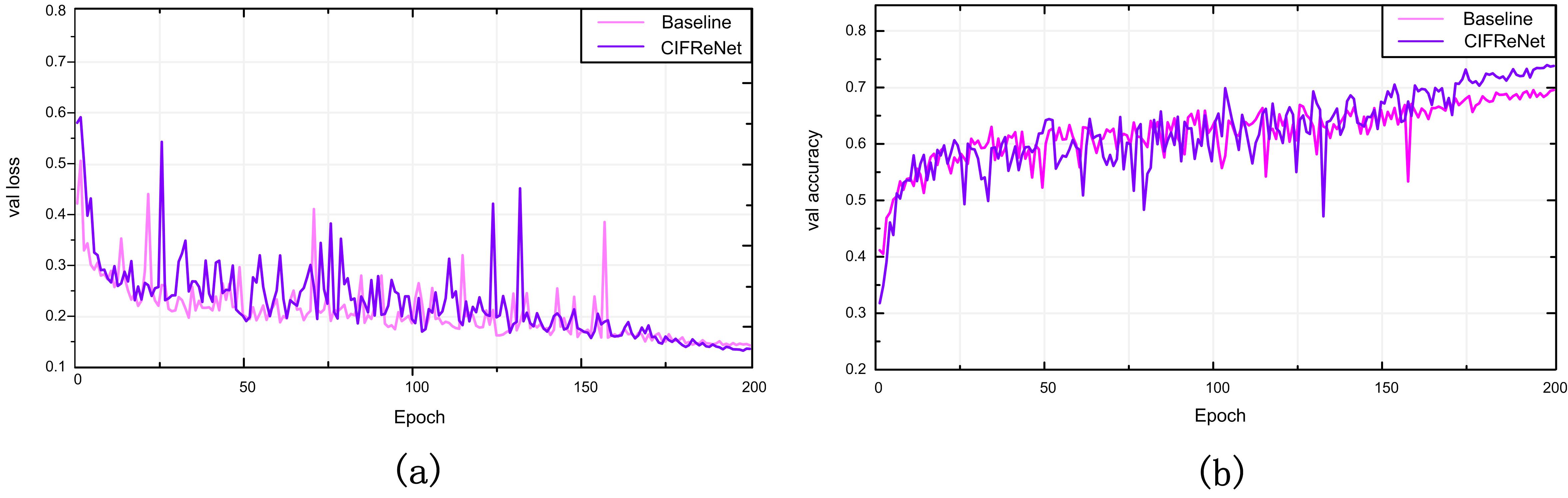}
% where an .eps filename suffix will be assumed under latex,
% and a .pdf suffix will be assumed for pdflatex; or what has been declared
% via \DeclareGraphicsExtensions.
\caption{The validation loss curve and accuracy curve of CIFReNet during 200 training epochs on CityScapes validation set. (a) Validation loss vs. Epoch. (b) Validation accuracy vs. Epoch.}
\label{11}
\end{figure}

Finally, we present the comparison of the baseline network and CIFReNet in terms of loss and accuracy curves on CityScapes validation set, as illustrated in Fig. \ref{11}(a) and Fig. \ref{11}(b). It reveals that CIFReNet could achieve a better convergence and higher accuracy than the baseline network. As shown in Table \ref{XI}, CIFReNet outperforms the baseline network by achieving a total of 5.11\% MIoU promotion. In summary, both qualitative and quantitative results verify that the proposed two modules can effectively enhance the network learning capacity.

\subsection{Evaluation on Cityscapes dataset}
Besides the above ablation studies, we also compare CIFReNet with other state-of-the-art methods on Cityscapes test set, as displayed in Table \ref{XII}. ``$\ast$" represents the results evaluated on Cityscapes validation set. Execution Speed (ES) is measured on NVIDIA TITAN X GPU cards. Methods marked with ``$\dagger$" represents the ES measured on other GPU cards. ``-" indicates that prior works have not published the corresponding value.

\begin{figure}[!t]
\centering
\includegraphics[width=3.4in]{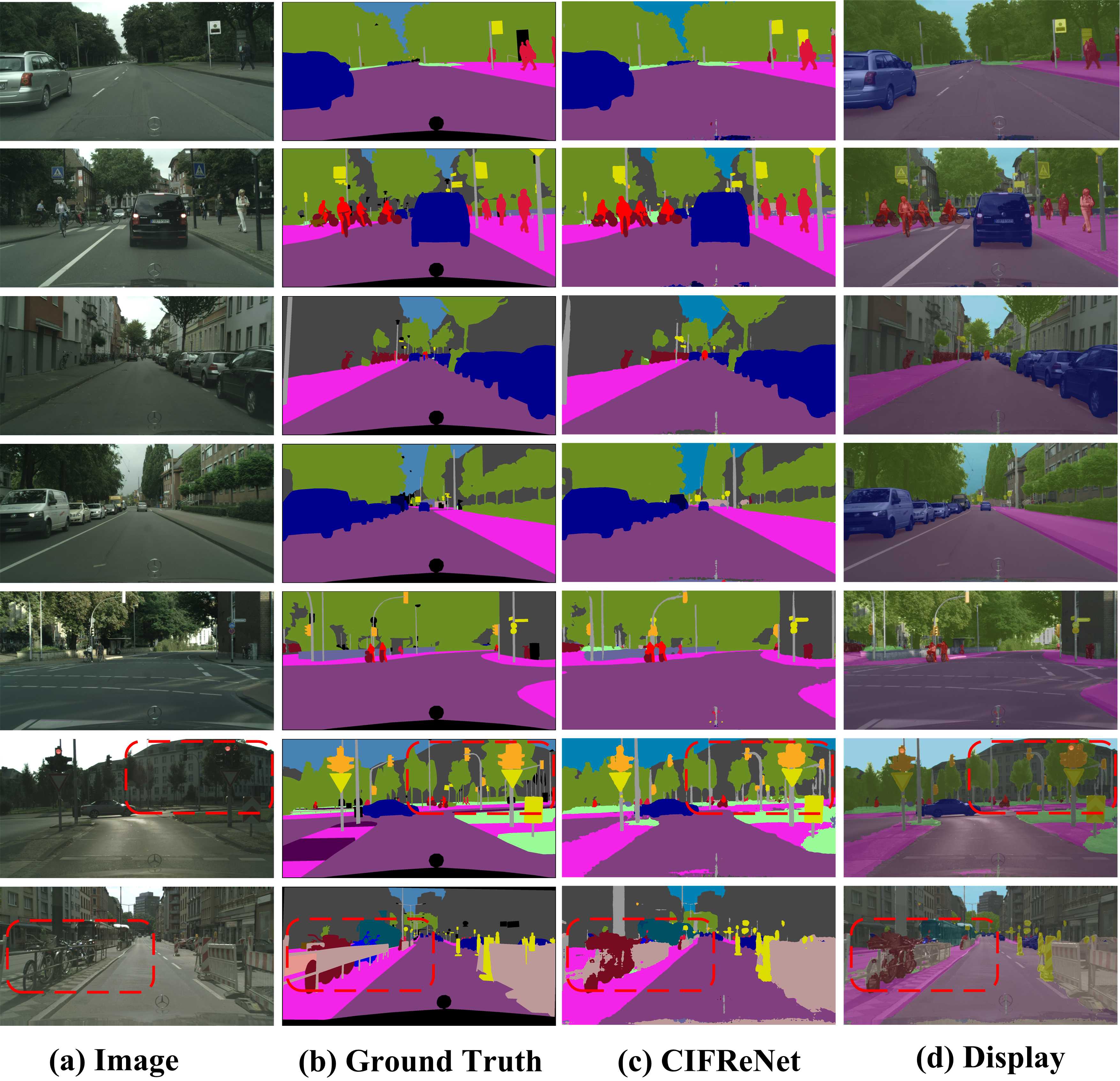}
% where an .eps filename suffix will be assumed under latex,
% and a .pdf suffix will be assumed for pdflatex; or what has been declared
% via \DeclareGraphicsExtensions.
\caption{Qualitative results on Cityscapes validation dataset when employing our best model (best viewed in color). The
last two rows show some failure cases.}
\label{12}
\end{figure}
\subsubsection{Performance Comparisons with Efficiency-oriented Methods}
Previous lightweight methods such as SegNet \cite{8}, LinkNet \cite{10}, and SQ \cite{14} have accelerated with great pace in speed, while failing to provide an accurate scene description for object parsing. Besides, ENet \cite{16} and Skip-MobileNet \cite{34} achieve excellent performance on model efficiency (e.g., execution speed, memory footprint or computational complexity). However, these methods only pursue efficiency by heavily compressing model, sacrificing too much segmentation accuracy. In contrast, the proposed CIFReNet, which obtains 70.9\% MIoU and yields a real-time speed of 62.5 FPS on a 640 $\times$ 360 resolution image, makes an obvious performance improvement on both accuracy and efficiency.
\begin{table}[!t]\caption{Performance comparisons on CamVid test set.}
\centering
  \begin{tabular}{c|c|c}
  \toprule[1pt]
  Module&$\sharp$Params(M)&MIoU(\%)\\
  \midrule[0.5pt]
  SegNet \cite{8}&29.5&46.4\\
  DeconvNet \cite{74}&252.0&48.9\\
  ENet \cite{16}&0.4&51.3\\
  LinkNet \cite{10}&11.5&55.8\\
  FCN8s \cite{29}&134.5&57.0\\
  Skip-MobileNet \cite{34}&3.4&58.8\\
  FC-DenseNet56 \cite{75}&1.5&58.9\\
  DeepLab-LFOV \cite{76}&37.3&61.6\\
  Dilation10 \cite{69}&140.8&65.3\\
  \midrule[0.5pt]
  CIFReNet(Ours)&1.9&64.5\\
  \bottomrule[1pt]
  \end{tabular}
\label{XIII}
\end{table}

Recent lightweight methods (e.g., GUNet \cite{18}, LW-RefineNet \cite{22}, ERFNet \cite{23}, ICNet \cite{24}, and L-DenseNet121 \cite{66}) have made a good trade-off between accuracy and speed, but deploying these models on edge devices is difficult due to heavy memory footprint or computational complexity. Differently, CIFReNet takes full advantages of compression techniques to relieve the resource burden, which is more lightweight and resource-saving. Particularly, the number of parameters and computational complexity are significantly reduced to 1.9 M and 7.3 GFLOPs. Besides, CIFReNet outperforms almost all methods stated above in terms of accuracy.

\subsubsection{Performance Comparisons with Accuracy-oriented Methods}
Table \ref{XII} also displays the results of CIFReNet compared with high-accuracy methods. We can observe that PSPNet \cite{4}, RefineNet \cite{9}, and Dilation10 \cite{69} employ costly VGGNets-like or ResNets-like base networks, which take more than 2 seconds for model inference. In contrast, CIFReNet achieves a significant progress in both speed and computation cost when dealing with a high resolution image, and obtains similar or slightly better accuracy compared with most of them (e.g., DeepLab V2 \cite{5}, DLC \cite{72}, and FRRNet \cite{73}). Without any post-processing operations (e.g., Dense CRF \cite{5}), the proposed CIFReNet reaches a better trade-off among overall performance. Fig. \ref{12} presents some visual examples on Cityscapes validation set. As seen, CIFReNet suffers from commonly challenging issues such as smooth boundary or obscure objects, which we would deal with in future works.

\begin{figure}[!t]
\centering
\includegraphics[width=3.3in]{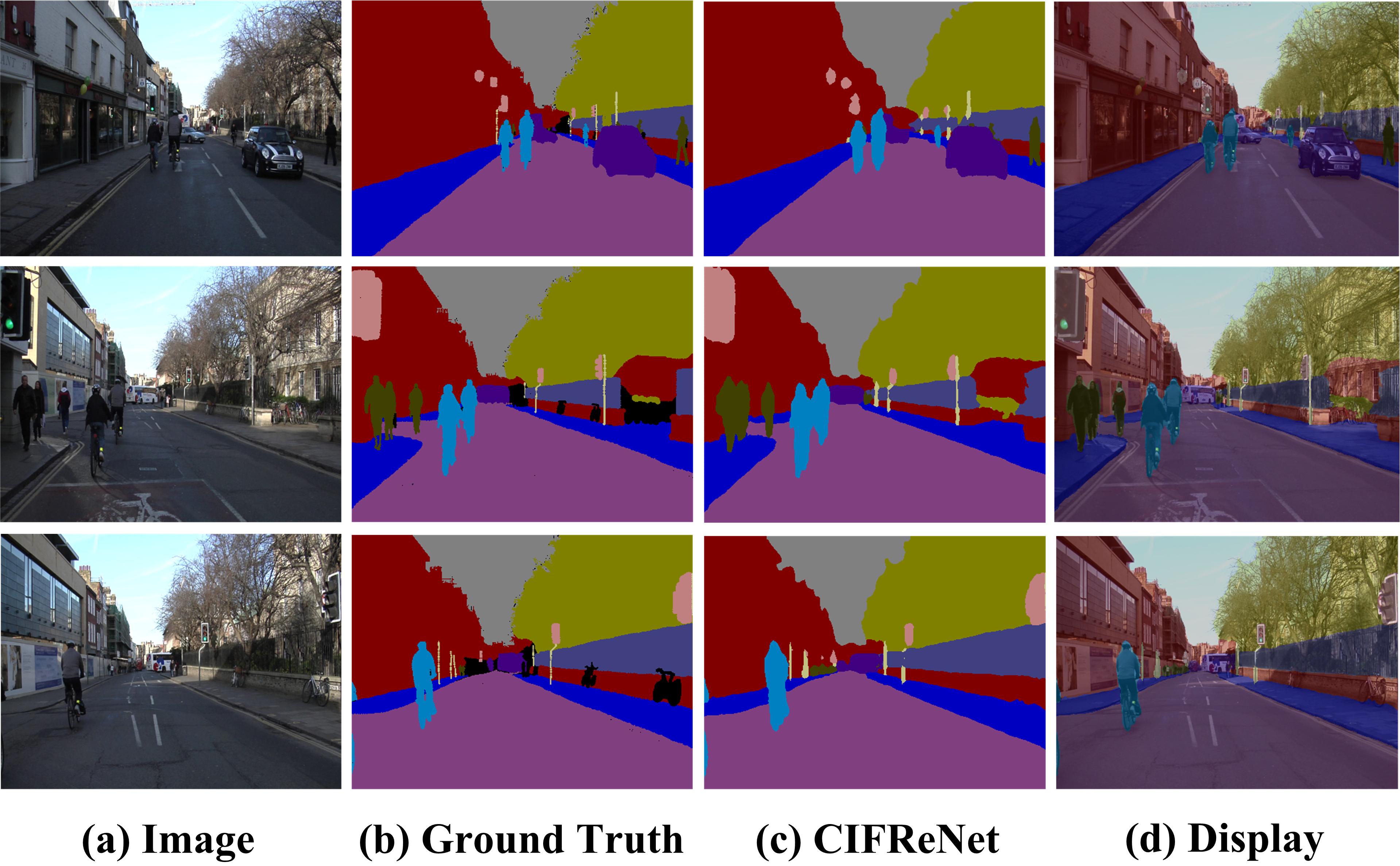}
% where an .eps filename suffix will be assumed under latex,
% and a .pdf suffix will be assumed for pdflatex; or what has been declared
% via \DeclareGraphicsExtensions.
\caption{Qualitative results on CamVid validation dataset when employing our best model (best viewed in color).}
\label{13}
\end{figure}

\begin{table}[!t]\caption{Performance comparisons on Helen test set.}
\centering
  \begin{tabular}{c|c|c}
  \toprule[1pt]
  Module&$\sharp$Params(M)&MIoU(\%)\\
  \midrule[0.5pt]
  FCN8s \cite{29}&134.5&40.7\\
  ENet \cite{16}&0.4&48.2\\
  SegNet \cite{8}&29.5&56.7\\
  ESPNet \cite{17}&0.2&60.7\\
  LinkNet \cite{10}&11.5&63.0\\
  CGNet \cite{77}&0.5&66.3\\
  DeepLab-LFOV \cite{76}&37.3&69.4\\
  ERFNet \cite{23}&2.1&70.1\\
  DeepLab V2 \cite{5}&44.0&70.9\\
  \midrule[0.5pt]
  CIFReNet(Ours)&1.9&71.3\\
  \bottomrule[1pt]
  \end{tabular}
\label{XIV}
\end{table}
\subsection{Evaluation on CamVid dataset}
We conduct another experiment on CamVid test set to further evaluate the performance of CIFReNet. The image resolution in CamVid dataset is much smaller than the one in Cityscapes dataset, so we set the dilation rates \textit{D$_{s}$}, \textit{D$_{m}$}, and \textit{D$_{l}$} of three DSP blocks to \{1, 2, 3, 5\}, \{5, 7, 9, 11\}, and \{11, 13, 15, 17\}, in order to effectively gather information from low-resolution feature maps. As reported in Table \ref{XIII}, CIFReNet achieves 64.5\% MIoU with only 1.9 M parameters, which reaches a better trade-off between accuracy and efficiency than accuracy-oriented (e.g., Dilation10 \cite{69}, DeepLab-LFOV \cite{76}) methods or efficiency-oriented (e.g., SegNet \cite{8}, ENet \cite{16}, and Skip-MobileNet \cite{34}). Some qualitative results on CamVid validation set are shown in Fig. \ref{13}.

\begin{figure}[!t]
\centering
\includegraphics[width=3.2in]{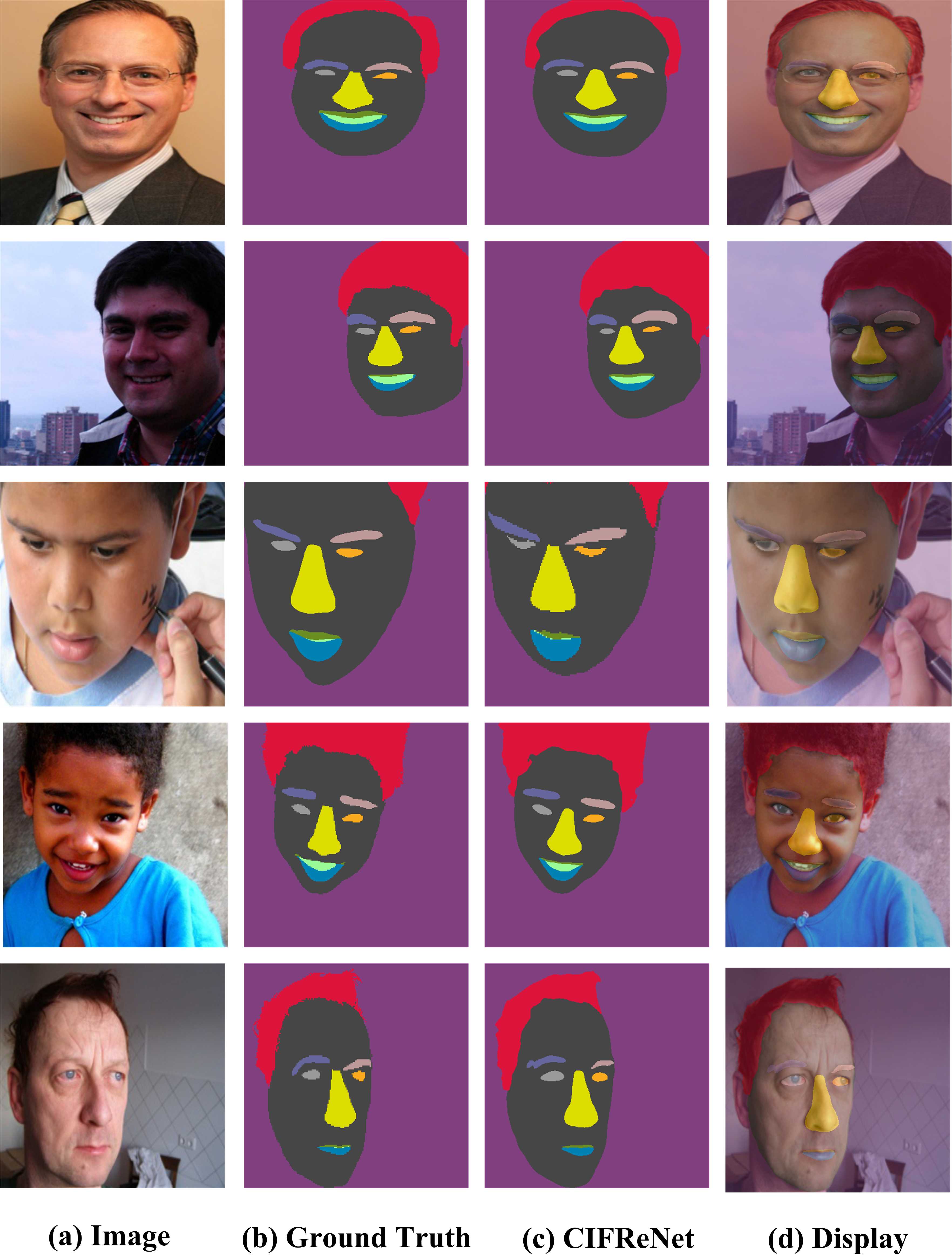}
% where an .eps filename suffix will be assumed under latex,
% and a .pdf suffix will be assumed for pdflatex; or what has been declared
% via \DeclareGraphicsExtensions.
\caption{Qualitative results on Helen test dataset when employing our best model (best viewed in color).}
\label{14}
\end{figure}
\subsection{Evaluation on Helen dataset}
 In order to verify the generalization of CIFReNet, we carry out some additional experiments to compare our method with current state-of-the-art methods on Helen test set. We observe that the size of image resolution in Helen dataset is generally about 400 $\times$ 400, so the value of \textit{D$_{s}$}, \textit{D$_{m}$} and \textit{D$_{l}$} of three DSP blocks are set to \{1, 2, 3, 5\}, \{3, 5, 7, 11\}, and \{7, 9, 11, 13\} to encode more useful contextual information. Semantic segmentation performance are specifically reported in Table \ref{XIV}. As seen, CIFReNet achieves 71.3\% MIoU with only 1.9 M parameters, which outperforms all methods mentioned above. These results indicate that our method can also make a promising trade-off between accuracy and efficiency for facial object parsing. Fig. \ref{14} presents some prediction visualizations of CIFReNet.

\section{Conclusion}
In this paper, we propose a lightweight Context-Integrated and Feature-Refined Network (CIFReNet) for object parsing. Specifically, our method consists of two core components: Long-skip Refinement Module (LRM) and Multi-scale Contexts Integration Module (MCIM). The LRM is designed to provide a highway for low-frequency information learning and boost the feature refinement efficiently. The MCIM utilizes Dense Semantic Pyramid (DSP) blocks and global features to encode diverse contextual information and obtain larger field of view in an economical way. Experiments show that the proposed CIFReNet not only achieves precise segmentation results but also relieves the burden of model efficiency, which makes it of great potentiality for deployment on resource-constrained intelligent devices.

In future works, we will make further discussions about two aspects: \textit{$\textbf{a}$}) Regardless of the dataset size, a scale-adaptive model will be designed to achieve steady performance at a faster speed. \textit{$\textbf{b}$}) The issue on how to effectively enhance the performance of boundary prediction will be further explored, which is supposed to replace previous costly measures.

\section*{Acknowledgment}
This work was supported in part by National Natural Science Foundation of China under Grant No. 61702176 and Hunan Provincial Natural Science Foundation of China under Grant No.2017JJ3038.
% or

%\appendices
%\section{XXX}
\ifCLASSOPTIONcaptionsoff
  \newpage
\fi

\end{document}